\documentclass[acmsmall,screen]{acmart}

\setcopyright{acmcopyright}
\copyrightyear{2021}
\acmYear{2021}
\acmDOI{10.1145/3387165}

\acmJournal{TiiS}
\acmVolume{11}
\acmNumber{3--4}
\acmArticle{22}
\acmMonth{12}

\usepackage{csquotes}
\usepackage{array}
\usepackage[export]{adjustbox}
\usepackage{subcaption}

\makeatletter
\g@addto@macro{\UrlBreaks}{\UrlOrds}
\makeatother

\newcommand{\tsnet}{\textit{t}}

\makeatletter
\newcommand{\pattern}[2][]{%
  \textsf{\textbf{P#1}}%
  \@bsphack
  \csname phantomsection\endcsname 
  \def\@currentlabel{\textsf{\textbf{P#1}}}{#2}%
  \@esphack
}
\makeatother

\newcommand{\PXspace}{%
    \hskip .07em plus .05em minus .03em}
\newcommand{\pathexplorer}{\texorpdfstring%
    {Pro\-jec\-tion\PXspace{}Path\PXspace{}Ex\-plor\-er}%
    {ProjectionPathExplorer}%
    }

\begin{document}

\title{\pathexplorer: Exploring Visual Patterns in Projected Decision-Making Paths}

\titlenote{This paper is an extended version of the workshop paper \enquote{Visualization of Rubik's Cube Solution Algorithms}~\cite{steinparz_visualization_2019}.}



\author{Andreas Hinterreiter}
\affiliation{%
  \institution{Johannes Kepler University Linz}
  \streetaddress{Altenberger Straße 69}
  \city{Linz}
  \country{Austria}
  \postcode{4040}
}
\email{andreas.hinterreiter@jku.at}
\affiliation{%
  \institution{Imperial College London}
  \streetaddress{South Kensington Campus}
  \city{London}
  \country{UK}
  \postcode{SW7 2AZ}
}
\email{a.hinterreiter@imperial.ac.at}

\author{Christian Steinparz}
\affiliation{%
  \institution{Johannes Kepler University Linz}
  \streetaddress{Altenberger Straße 69}
  \city{Linz}
  \country{Austria}
  \postcode{4040}
}
\email{christian.steinparz@jku.at}

\author{Moritz Sch\"ofl}
\affiliation{%
  \institution{Johannes Kepler University Linz}
  \streetaddress{Altenberger Straße 69}
  \city{Linz}
  \country{Austria}
  \postcode{4040}
}
\email{moritz.schoefl@jku.at}

\author{Holger Stitz}
\affiliation{%
  \institution{Johannes Kepler University Linz}
  \streetaddress{Altenberger Straße 69}
  \city{Linz}
  \country{Austria}
  \postcode{4040}
}
\email{holger.stitz@jku.at}
\affiliation{%
  \institution{datavisyn GmbH}
  \streetaddress{Altenberger Straße 69}
  \city{Linz}
  \country{Austria}
  \postcode{4040}
}
\email{holger.stitz@datavisyn.io}

\author{Marc Streit}
\affiliation{%
  \institution{Johannes Kepler University Linz}
  \streetaddress{Altenberger Straße 69}
  \city{Linz}
  \country{Austria}
  \postcode{4040}
}
\email{marc.streit@jku.at}

\renewcommand{\shortauthors}{A.~Hinterreiter, C.~Steinparz, M.~Sch\"ofl, H.~Stitz, and M.~Streit}

\begin{abstract}
    In problem-solving, a path towards solutions can be viewed as a sequence of decisions.
    The decisions, made by humans or computers, describe a trajectory through a high-dimensional representation space of the problem.
    By means of dimensionality reduction, these trajectories can be visualized in lower-dimensional space.
    Such embedded trajectories have previously been applied to a wide variety of data, but analysis has focused almost exclusively on the self-similarity of single trajectories.
    In contrast, we describe patterns emerging from drawing many trajectories---for different initial conditions, end states, and solution strategies---in the same embedding space.
    We argue that general statements about the problem-solving tasks and solving strategies can be made by interpreting these patterns.
    We explore and characterize such patterns in trajectories resulting from human and machine-made decisions in a variety of application domains: logic puzzles (Rubik's cube), strategy games (chess), and optimization problems (neural network training).
    We also discuss the importance of suitably chosen representation spaces and similarity metrics for the embedding.
\end{abstract}




\begin{teaserfigure}
    \includegraphics[width=\textwidth]{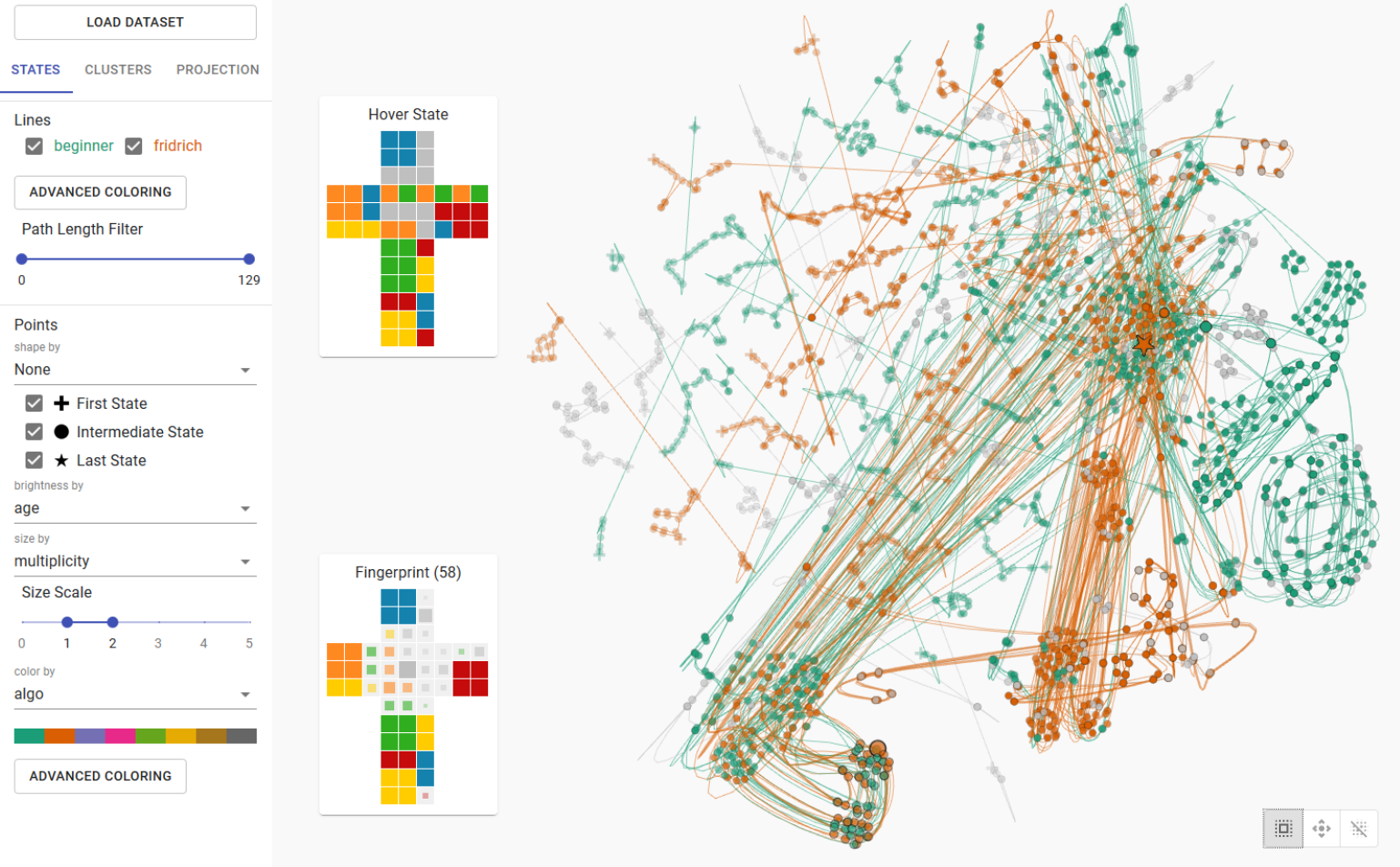}
    \caption{The \pathexplorer{} visualization prototype allows exploration of patterns in decision-making paths. Multiple series of high dimensional states are visualized as trajectories through a joint embedding space.}
    \Description{User interface of the \pathexplorer{} prototype.} 
    \label{fig:teaser}
\end{teaserfigure}

\maketitle

\section{Introduction}

The act of solving problems usually involves sequences of decisions.
In most contexts, decisions can be viewed as transitions between states in a  representation space.
For example, solving logic puzzles such as the Rubik's cube requires decisions about how to transform an object from a random initial configuration to a solved final state.
Strategy games can also be viewed as  progressions through game states; each decision transforms the board state and leads closer to a player's victory or defeat.
Even optimization processes, such as the training of neural networks, involve decisions about transitioning from one intermediate state to anotherin the hope that the process ultimately converges to a final solution.

In many application domains, the decisions determining a transition from one state to another are  made consciously by human agents, often after a complex reasoning process.
In other cases, the transitions between states can be based on decisions made automatically by algorithms.
Examining patterns in the paths towards a problem solution can lead to a better understanding of approaches used by humans and machines to solve complex tasks.

For almost all real-world problems, the possible representation spaces are not only vast, but also high-dimensional.
Many attributes are needed to accurately describe each intermediate state along the pathway to a solution.
The high dimensionality of the state spaces makes assessing patterns in solution paths challenging.
Previously, dimensionality reduction techniques have been used in conjunction with trajectory visualizations to explore high-dimensional sequences of states~\cite{bach_time_2016}.

Research  has so far  focused mostly on discovering patterns in single solution paths.
However, the exact path towards a solution may depend strongly on the initial state for that particular instance of the problem.
For example, in the case of optimization algorithms, paths of convergence might depend on the chosen initialization.
In other domains, for instance in strategy games such as chess, the initialization is always the same.
Here, early decisions might force the solution path along one particular of many different tracks.

In both cases, it is not enough to view single solution paths without the context of many other paths.
In fact, many interesting patterns emerge only from \emph{multiple} paths.
In this paper, we present an extension of established trajectory techniques for visualizing high-dimensional paths through complex representation spaces.
We identify patterns emerging from sets of projected solution trajectories.
Using interactive visualization, we enable users to find such patterns and explore their relationships with the underlying real-world processes.

We present a general interactive prototype for the visualization of multiple high-dimensional solution trajectories and show how minimal adaptations can make this prototype useful in a diverse range of application domains.
Specifically, we present examples in which we compare Rubik's cube solution algorithms, find patterns in chess games, and assess meaningful representations of neural network training processes.
These examples show how extending the self-similarity concept of single projected solution trajectories to include similarities between multiple trajectories can provide more insights into problem-solving processes.

This paper is structured as follows.
Section~\ref{sec:related-work} gives an overview of publications related to our work.
In Section~\ref{sec:technique} we introduce our technique and discuss how it extends existing work.
In Section ~\ref{sec:patterns} we characterize the possible patterns emerging from our visualization technique.
Section~\ref{sec:prototype} is a description of our prototype implementation of \pathexplorer{}, which we used to analyze data from several application domains.
In Section~\ref{sec:applications} we show the results of these analyses.
The general insights gained from these applications and ideas for future work are discussed in Section~\ref{sec:discussion}.
Section~\ref{sec:conclusion} concludes the paper.

\section{Related Work}
\label{sec:related-work}

We study patterns emerging from multiple dimensionality-reduced trajectories through a problem state space.
First we discuss the general importance of trajectories for time series visualization.
We then briefly review the most widely used dimensionality reduction techniques.
Finally, and most pertinently to our approach, we discuss previous attempts at using trajectories for visualizing dimensionality-reduced data.

\subsection{Trajectories for Time Series Visualization}

Visualizing the progression of processes (algorithms, games, etc.) through different states is equivalent to visualizing multivariate time series data.
A dataset with two quantitative attributes that change over time can be readily visualized as a trajectory through 2D data space, with time varying between the support points.
The DimpVis system~\cite{kondo_dimpvis:_2014}, for instance, demonstrated the use of trajectories as an important factor in interactive displays of time series data.

Trajectories naturally play an important role in visual analytics of movement~\cite{andrienko_visual_2013}.
When many trajectories (also known as trails) are shown at once, visual clutter can be a problem.
To address this issue, edge-bundling techniques~\cite{holten_force-directed_2009,ersoy_skeleton-based_2011} have been adapted to trail visualization~\cite{peysakhovich_attribute-driven_2015,hurter_bundled_2013,du_trajectory_2015}.
In order to gain additional insights from spatial trajectories, numerous trajectory data mining techniques have been developed~\cite{zheng_trajectory_2015}.

\subsection{Similarity of Time Series}

Assessing the similarity between multiple time series has been studied extensively for various encodings of one-dimensional temporal data.
This includes line charts, heatmaps/colorfields and horizon graphs~\cite{gogolou_comparing_2019}, and image encodings~\cite{wang_encoding_2015}.

However, these encodings of one-dimensional data cannot exhibit many of the interesting patterns that arise from trajectories.
Such patterns are cycles, parallel segments, and/or points visited multiple times, and they result from the fact that---in case of the trajectory encoding---time varies implicitly along the path through multiple dimensions.
From these patterns a notion of self-similarity follows, which is arguably one of the strengths of the trajectory encoding.
Plotting multiple trajectories together enables assessment of both similarity \emph{between} time series and \emph{self}-similarity of individual time series.

\subsection{Dimensionality Reduction}

Often the data to be visualized as trajectories has more than two dimensions.
While some data attributes can be encoded in additional channels (e.g., line color, line width, or marker shape), dimensionality reduction becomes inevitable for 5 or more data dimensions.
A~multitude of dimensionality reduction techniques has been developed~\cite{van_der_maaten_dimensionality_2009}, and their usefulness for data visualization has been studied~\cite{engel_survey_2012}.
Principal component analysis (PCA) and multidimensional scaling (MDS) are two classic techniques used for dimensionality reduction.
In the 2000s, Isomap~\cite{tenenbaum_global_2000} and \tsnet-distributed stochastic neighbor embedding (\tsnet-SNE)~\cite{van_der_maaten_visualizing_2008} were presented---two non-linear techniques specifically created for data visualization.
Around the same time, deep autoencoder networks proved to be capable of learning low-dimensional codes of high-dimensional data~\cite{hinton_reducing_2006}.
More recently, uniform manifold approximation and projection (UMAP) has ben introduced as an alternative to \tsnet-SNE~\cite{mcinnes_umap:_2018}.
In the last years, projection techniques have been used increasingly for visualizations in explainable AI~\cite{hohman_visual_2018}.
Rauber et al., for instance, used \tsnet-SNE for visualizing how the activity of neural networks evolves throughout training~\cite{rauber_visualizing_2017}.
A~recent survey by Espadoto et al.~\cite{espadoto_towards_2019} compares over 40 different dimensionality reduction techniques quantitatively.

\subsection{Combining Dimensionality Reduction and Trajectories}

As stated above, without any additional encoding, trajectories can be readily used only to visualize time series with two attributes.
For time series with more attributes, a combination of dimensionality reduction with trajectory visualization can be a powerful approach.

Schreck et al.~used self-organizing maps (SOMs) to display projected trajectories of high-dimensional, time-resolved financial data~\cite{schreck_trajectory-based_2007}.
For document visualization, Mao et al.~derived representations of n-gram data akin to phase-space trajectories used in theoretical physics~\cite{mao_sequential_2007}.
Ward and Guo generalized the notion of n-grams to non-textual data, yielding a shape-space representation of snippets of time series data~\cite{ward_visual_2011}.
In TimeSeriesPath, data points of multivariate time series are projected using PCA, and shown as trajectories in 2D space~\cite{bernard_timeseriespaths:_2012}.
Schreck et al.~used self organizing maps (SOMs) to display projected trajectories of high-dimensional, time-resolved financial data~\cite{schreck_trajectory-based_2007}.
In ThermalPlot~\cite{stitz_thermalplot:_2016}, multi-attribute time-series data is reduced to a single point by means a user-defined degree-of-interest (DOI) function and plotted in a DOI space.
Items with similar temporal development are clustered and show similar trajectories.

All these similar approaches are summarized in the Time Curve idiom by Bach et al.~~\cite{bach_time_2016}. 
Time Curves are \enquote{based on the metaphor of folding a timeline visualization into itself so as to bring similar time points close to each other~\cite{bach_time_2016}}.
The versatility of the Time Curves technique is reflected in the diverse range of application domains: high-dimensional dynamic networks~\cite{van_den_elzen_reducing_2016,boz_visual_2019}, neural network learning~\cite{rauber_visualizing_2017}, user-interaction data~\cite{brown_modelspace:_2018}, trend and outlier detection~\cite{cakmak_time_2018}, sport-game visualization~\cite{zhu_performance_2016}, and temporal patterns in representation data~\cite{he_visualizing_2017}.

One example application in the Time Curves paper is the visualization of document edit histories~\cite{bach_time_2016}.
In this domain, similar approaches had been used previously, by treating sequences of n-grams similarly to multivariate time series.
Using PCA and differentiation, Mao et al.~derived visualizations of n-gram data akin to phase space trajectories used in theoretical physics~\cite{mao_sequential_2007}.
Ward and Guo generalized the notion of n-grams to non-textual data, yielding a shape-space representation of snippets of time series data~\cite{ward_visual_2011}.
The shape-space representation was, in turn, projected by PCA and visualized as trajectories in two dimensions.%

In an approach very similar to Time Curves, van den Elzen et al.~used \tsnet-SNE, among other projection techniques, to visualize high-dimensional states of dynamic networks~\cite{van_den_elzen_reducing_2016}.
A~recent Master's thesis~\cite{boz_visual_2019} features similar visualizations of network states.

Rauber et al.~\cite{rauber_visualizing_2017}~used \tsnet-SNE to visualize the activations in neural networks over time.
In ModelSpace, Brown et al.~\cite{brown_modelspace:_2018} visualize user interaction data as projected trajectories, aiming to reveal patterns in the users' intentions.
Cakmak et al.~used a \tsnet-SNE visualization for trend and outlier detection in projected time series~\cite{cakmak_time_2018}.
Zhu and Chen created dimensionality-reduced trajectories for basketball games, making use of landmark MDS~\cite{zhu_performance_2016}.

Finally, He and Chen used a time-aware implementation of \tsnet-SNE to visualize temporal patterns in representation data~\cite{he_visualizing_2017}.
This temporal \tsnet-SNE variant places similar high dimensional states at large distances in the embedding spaces, if the are temporally far away from each other.

\subsection{Context of This Work}

Among these related works, the Time Curves paper by Bach et al.~\cite{bach_time_2016}is perhaps the most relevant one for our work.
The usefulness of the Time Curves idiom is based on a number of characteristics of the projected trajectories%
, such as point density and irregularity, as well as patterns such as transitions, cycles, and oscillations~\cite{bach_time_2016}.
However, Bach et al.~focused on the interpretation of single trajectories and only hinted at the power of constructing and interpreting multiple trajectories together.
We build upon their classification, but focus on additional patterns emerging from \emph{multiple} trajectories.

Other authors have demonstrated the potential of constructing multiple trajectories in the same embedding space in particular contexts.
Zhu and Chen were able to detect outliers  among multiple basketball games~\cite{zhu_performance_2016}.
Brown et al.~discovered differences in velocities along multiple trajectories from interaction data, and traced back these differences to varying user speeds~\cite{brown_modelspace:_2018}.
However, little work has focused on identifying general patterns emerging from multiple embedded trajectories throughout different application domains.

In this paper we apply our approach to data from diverse application domains, including games (chess), puzzles (Rubik's cube), and AI (neural network training).
Depending on the application context, the number of trajectories we plot conjointly in the same embedding space varies between a few and several hundred.
We argue that general statements about the visualized decision-making processes can be made based on similarities between multiple trajectories.

\section{Technique}
\label{sec:technique}

Prior to visualizing paths of problem solutions with trajectories through an embedding space, a number of important decisions have to be made.
The visualization depends strongly on how the real-world states are represented as vectors.
The distance metric for these vectors needs to be chosen carefully, as must the dimensionality reduction technique.
Finally, a suitable visual encoding for the low-dimensional states and trajectories needs to be chosen.
We discuss all these aspects in the following section.
We also present a discussion of possible patterns emerging from multiple trajectories in the same embedding space.

\subsection{State-Space Representation}
\label{sec:prelim}

As previously mentioned, our visualization approach is similar to  the Time Curve technique introduced by Bach et al.~\cite{bach_time_2016}.
Time Curves are trajectories drawn through projections of high-dimensional data points.
They are constructed by defining a distance metric in the high-dimensional domain and embedding the data points in two dimensions such that certain characteristics of the high-dimensional distance metric are preserved.
The projected points are connected depending on the timestamps associated with each high-dimensional data point.
Table~\ref{tab:time-curve-definitions} lists the key terms and definitions in the context of dimensionality-reduced trajectories.

The timestamps of the projected points are not relevant to their placement in the embedding space.
Only the relative temporal ordering is used for connecting the embedded points.
In most scenarios, projected trajectories are thus visualizations of sequences (or ordered sets) of states.
This trajectory technique can therefore be applied to any process that involves an object undergoing several sequential states, and state indices can entirely replace actual timestamps.
Bach et al.~use the term \enquote{data snapshot} to refer to the individual high-dimensional data items that---together with the timestamps---make up the multivariate time series.
They call the pairs of timestamps and data snapshots \emph{time points}.
In contrast, we refer to data snapshots as \emph{states}, and we call the snapshot domain \emph{state space}.
This stresses our focus on sequential processes in which the actual time values are not necessarily of interest or even unavailable (e.g., as in the guiding example below).

An essential requirement for constructing a meaningful visualization of trajectories through an embedding space is the way in which the real-world object (that goes through the sequential steps) is \enquote{digitized}.
In our opinion, this is the first, and possibly most important decision when combining projection with trajectories, but it was not discussed in detail by Bach et al.
We refer to this digitization as \emph{state-space representation} and define it as a mapping from the \enquote{real world} \(\mathcal{R}\) to the state space \(\mathbb{S}\).
Importantly, the state-space representation directly influences the meaning of any potential distance metrics.

A state-space representation incorporates one of possibly many selections of an object’s attributes.
All data that is not included is potentially interesting \emph{metadata}, which is not used directly to calculate the trajectories' support points, but can be encoded additionally (e.g.,~by coloring trajectories or points categorically depending on the associated metadata).
Thus, the actual value of the timestamp \(t_i\) of a time point \((t_i, s_i)\) can also be treated as metadata.

\begin{table}
    \centering
    \caption{Key terms and definitions for the construction of Time Curves as given by Bach et al.~\cite{bach_time_2016}, and our own notations for state-space representations and dimensionality reduction.}
    \label{tab:time-curve-definitions}
    \begin{tabular}{ll}
        \toprule
        Notation                                    & Explanation       \\
        \midrule
        \(t_i \in \mathbb{R}\)                      & timestamp                     \\
        \(\mathbb{S}\)                              & snapshot domain (state space) \\
        \(s_i \in \mathbb{S}\)                      & data snapshot (state)         \\
        \(p_i = (t_i, s_i)\)                        & time point                    \\
        \(P = \{p_1, \dots, p_n\}\)                 & temporal dataset              \\
        \(d\colon \mathbb{S}^2 \to \mathbb{R}^+\)   & distance metric               \\
        \(D = [d_{ij}] = [d(p_i, p_j)]\)            & distance matrix               \\
        \(P_D = \{P, D\}\)                          & temporal similarity dataset   \\
        \midrule
        \(r\colon \mathcal{R} \to \mathbb{S}\)      & state-space representation    \\
        \(f\colon \mathbb{S} \to \mathbb{R}^2\) or
        \(f\colon \mathbb{S}^n \to (\mathbb{R}^2)^n\)& embedding*\\ 
        \bottomrule
        \multicolumn{2}{l}{\footnotesize * Domain and codomain of the embedding function depend on the technique chosen.}
    \end{tabular}
\end{table}

As a guiding example throughout this section, we apply our visualization approach to data from two well-known sorting algorithms: bubble sort and quicksort.
We first create all possible permutations of the list \((1,\dots,n)\) for a given length \(n\).
We apply each of the two algorithms to each permutation and record all intermediate steps.
For visualizing the progression of the algorithms, we first need to convert the list objects from \(\mathcal R\) to a suitably chosen state space \(\mathbb S\) by means of the state-space representation \(r\).
We decide that we want to compare lists in terms of \enquote{how shuffled} they are.
This similarity can be measured using an edit distance.
There are now two approaches for obtaining meaningful distances:
\begin{itemize}
    \item We can directly use the symmetric group of all permutations of lists with length \(n\) as our state space:~\(\mathbb{S} = \mathfrak{S}_n\).
    This allows us to simply use an edit distance of our choice as metric \(d\), such as the Hamming distance or the Damerau--Levenshtein distance.
    \item Alternatively, we can use another representation of lists in which a distance of our choice (e.g., the squared euclidean distance) corresponds to an edit distance.
    In the case of our permuted lists, such a representation can be constructed by performing a one-hot encoding: \(\mathrm{enc}((x_1,\dots,x_n)) = \mathrm{flat}((\mathrm{sparse}_n(x_1),\dots,\mathrm{sparse}_n(x_n)))\).
    Here, \(\mathrm{flat}(M)\) flattens a matrix \(M\) to a vector.
    The vector \(\mathrm{sparse}_n(i) = (\delta_{1i},\dots,\delta_{ni})\) is a vector of length \(n\) in which only the \(i\)th entry is set to 1 and all other entries are set to 0.
    In the formula for \(\mathrm{sparse}_n(i)\), \(\delta\) represents the Kronecker delta.
    The resulting vectors have a length of \(n^2\), and the state space is \(\mathbb{S} = \{0,1\}^{(n^2)}\).
    In this state space, \(d(a,b) = \mathrm{Euclidean}(a,b)^2 / 2\) is exactly equivalent to the Hamming distance.
\end{itemize}

Since most dimensionality reduction techniques accept distance matrices as input, or can be readily adapted to accept non-numeric data, the first approach is usually preferable.
However, the second approach can help to construct meaningful distances via intermediate encodings, when the \enquote{true} distance is computationally infeasible.
This will become evident in the discussion of the distance metric we used to compare states of a Rubik's cube in Section~\ref{sec:rubik}.

\subsection{Dimensionality Reduction}

In this work, we focus on visualizing \emph{multiple} trajectories conjointly.
This also affects the calculation of the embedding used for dimensionality reduction.
In order to embed all trajectories in the same space, the individual states of all sequences must be projected in conjunction.
This means that identical elements in the high-dimensional state space are more likely to occur multiple times.
This is especially the case when the visualized real-world process is constrained to always start from the same initial state, always ends in the same final state, or has fixed intermediate states that must be traversed (see also the discussion of patterns in Section~\ref{sec:patterns}).
In our guiding example, each application of either of the two sorting algorithms terminates in the same state, namely the sorted array \((1,\dots,n)\).
This state will be present in our dataset a total of \(m\) times, where \(m\) is the number of different initial states multiplied by the number of sorting algorithms used.

In our experience, most implementations of \tsnet-SNE exhibit the following behavior:
when an input set \(X = \{x_1, x_2, \dots, x_n\}\), with some identical high-dimensional points \(x_i = x_j\), is embedded, \(y_i = y_j\) is not guaranteed to hold for the resulting embedding \(f(X) = \{y_1, y_2, \dots, y_n\}\).
Since the objective function of \tsnet-SNE is optimized for all points simultaneously, identical embedding coordinates for identical input values cannot be guaranteed---although the embedding coordinates will typically be very close together.
Depending on the projection technique, keeping duplicates in the input set can thus offset, and thereby highlight, points that are visited more often than others.
If exact identity is required, duplicates must be removed beforehand.

\begin{figure}
    \centering
    \begin{minipage}[b]{0.33\textwidth}
        \centering
        \includegraphics[width=\textwidth]{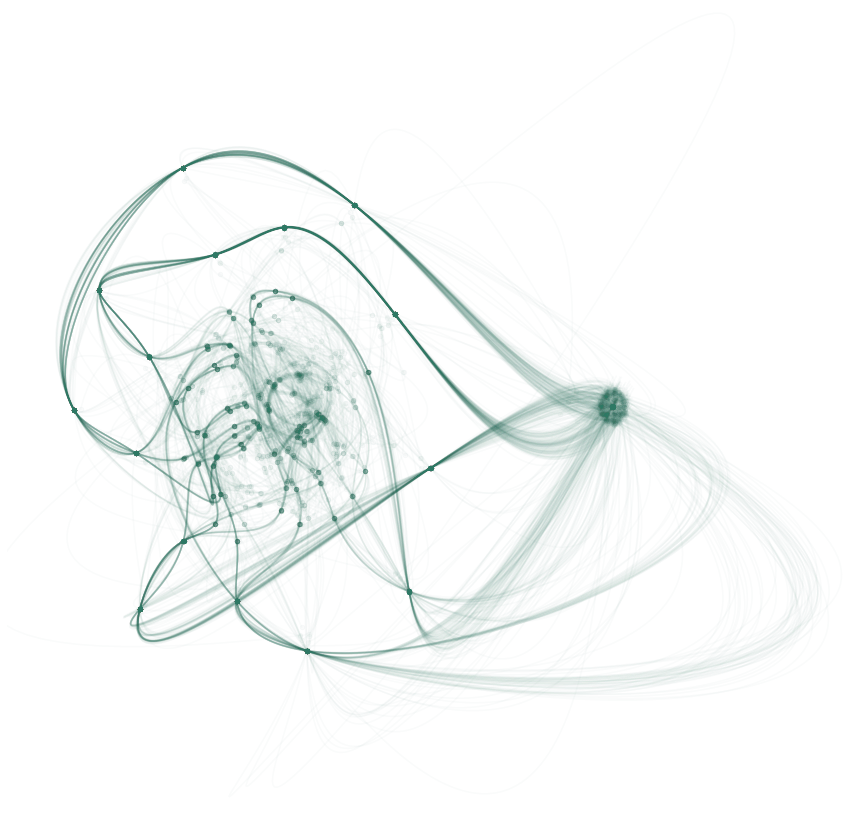}
        \subcaption{Bubble sort (\tsnet-SNE)}
    \end{minipage}%
    \begin{minipage}[b]{0.33\textwidth}
        \centering
        \includegraphics[width=\textwidth]{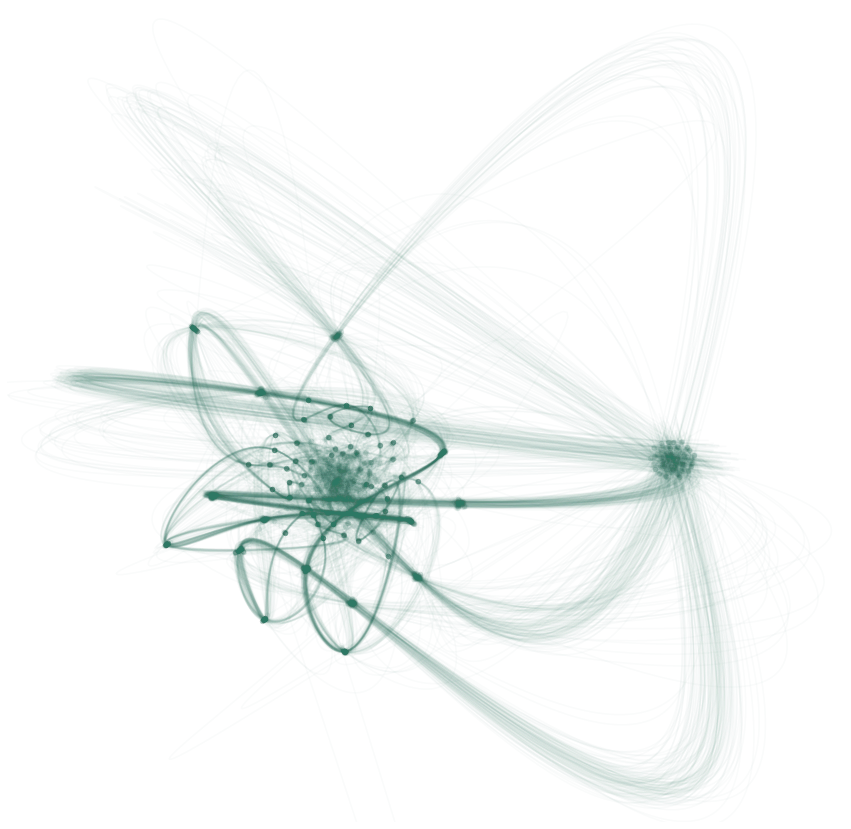}
        \subcaption{Bubble sort (UMAP)}
    \end{minipage}%
    \begin{minipage}[b]{0.3\textwidth}
        \centering
        \includegraphics[width=\textwidth]{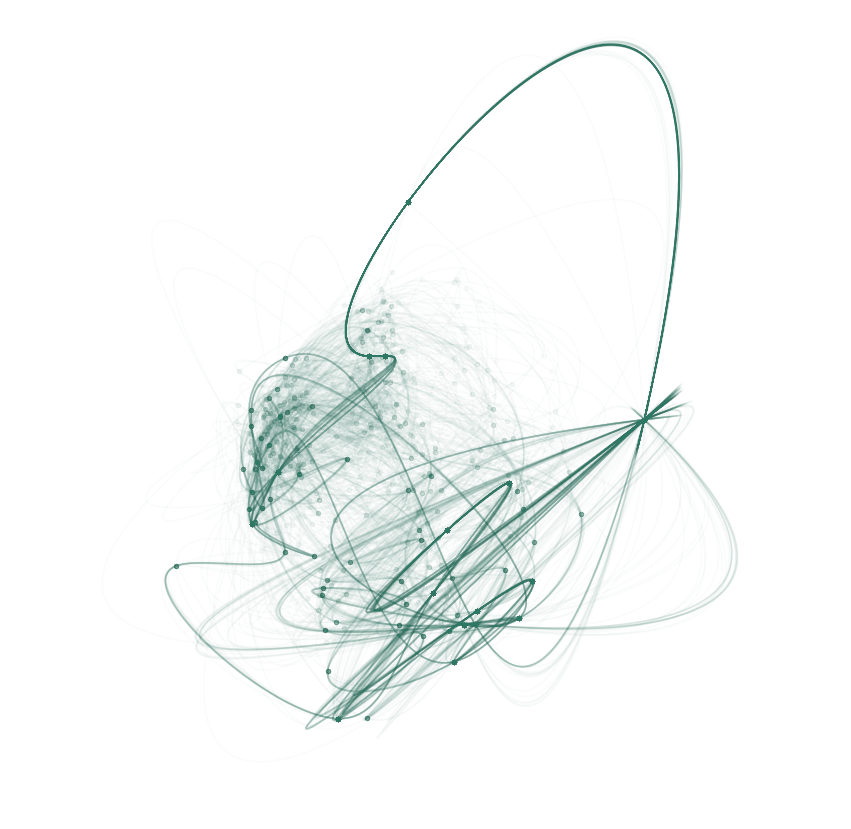}
        \subcaption{Bubble sort (Isomap)}
    \end{minipage}
    \medskip
    
    \begin{minipage}[b]{0.33\textwidth}
        \centering
        \includegraphics[width=\textwidth]{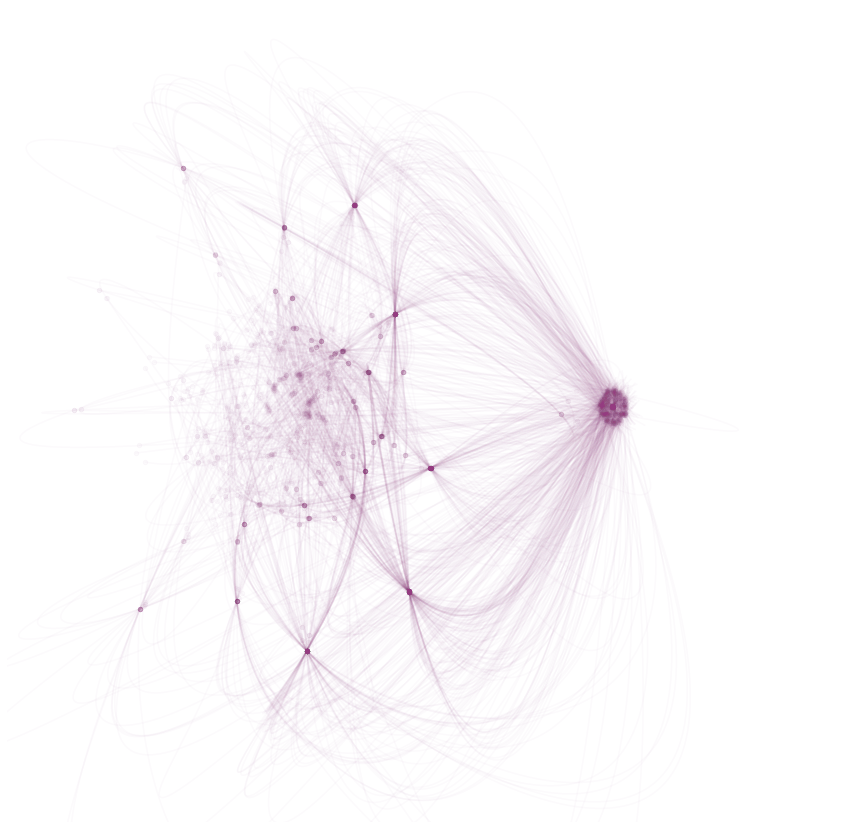}
        \subcaption{Quicksort (\tsnet-SNE)}
    \end{minipage}%
    \begin{minipage}[b]{0.33\textwidth}
        \centering
        \includegraphics[width=\textwidth]{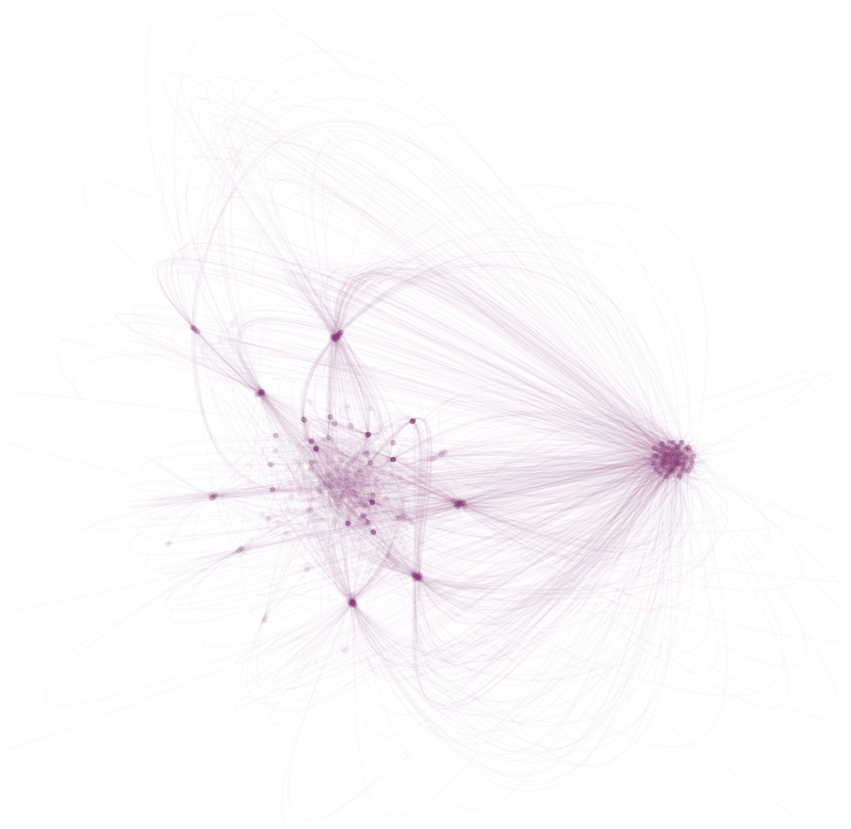}
        \subcaption{Quicksort (UMAP)}
    \end{minipage}%
    \begin{minipage}[b]{0.33\textwidth}
        \centering
        \includegraphics[width=\textwidth]{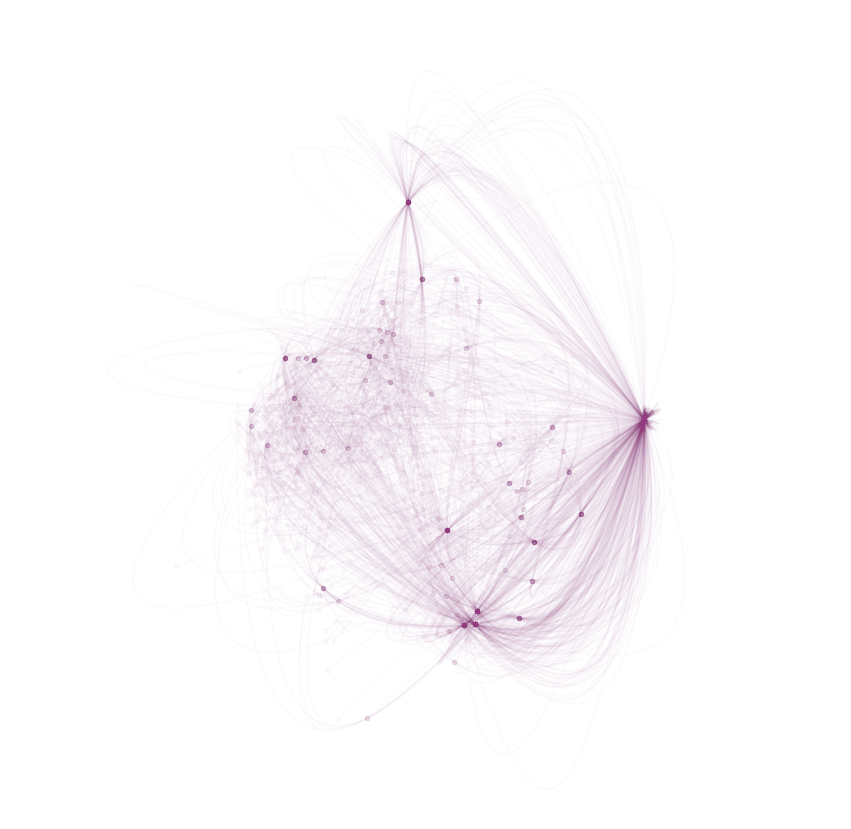}
        \subcaption{Quicksort (Isomap)}
    \end{minipage}
    \caption{Sorting trajectories for bubble sort~(a--c) and quicksort~(d--f), applied to all permutations of a list of length 6 with unique entries.
    Three different dimensionality reduction techniques are shown: (a,\,d)~\tsnet-SNE with a perplexity of 100 and an exaggeration factor of 2 during the main optimization; (b,\,e)~UMAP with 25 nearest neighbors and a minimum distance of 0.1; and (c,\,f)~Isomap with 75 nearest neighbors.
    For each projection technique, data for quicksort and bubble sort was combined, to obtain a shared embedding space.
    Duplicates were not removed prior to the embedding.
    For \tsnet-SNE and UMAP, the duplicate sorted arrays form a pronounced cluster, and the shared intermediate points are more salient.
    All plots were rotated such that the sorted end states are located to the right.
    }
    \Description{Comparison of different embedding techniques for trajectories of intermediate states of bubble sort and quicksort, respectively.}
    \label{fig:guiding-example}
\end{figure}

Figure~\ref{fig:guiding-example} presents the sorting trajectories of all permutations of the list \((1,2,3,4,5,6)\) for bubble sort and quicksort.
It shows results for projecting the states by means of \tsnet-SNE, UMAP, and Isomap.

To ensure comparability between the sorting algorithms, the results of bubble sort and quicksort were projected together into a shared embedding space by using one of the three dimensionality reduction techniques.
We optimized the hyperparameters of each technique to obtain visually appealing results.
Regardless of the embedding routine, the visualizations for bubble sort (Figure~\ref{fig:guiding-example}~(a--c)) exhibit long and winding trajectory bundles.
Many different sorting trajectories must go through the same few states before finally ending up in the sorted end state.
For quicksort (Figure~\ref{fig:guiding-example}~(d--f)), the paths from the initial states to the sorted state are generally shorter and do not overlap as much as in the case of bubble sort.
This reflects the fact that, on average, quicksort passes only 3.5 intermediate states from start to finish, while bubble sort takes 7.5 steps for the given lists of length 6.
Furthermore, bubble sort is limited to simple swapping actions, which forces many sorting trajectories to take similar paths through the state space.

For all three embedding techniques, we did not remove duplicates prior to the embedding.
For Isomap, duplicates in the input list make no difference, as duplicate high-dimensional points are mapped to to the same two-dimensional point.
For \tsnet-SNE and UMAP, however, high-dimensional duplicates are embedded as clusters in two dimensions.
This behavior of \tsnet-SNE and UMAP is especially relevant to the cluster of sorted end states near the right-hand side of the plots in Figure~\ref{fig:guiding-example}~(a), (b), (d), and (e).
The clusters make the trajectory bundles thicker and more easily traceable.

One insight from the comparison in Figure~\ref{fig:guiding-example} is that, with the right parameter choice, \tsnet-SNE and UMAP can yield very similar plots.
This is the case when the chose perplexity parameter of \tsnet-SNE is high enough for some global structure to be preserved.
A typical argument in favor of UMAP is that \tsnet-SNE becomes inefficient for high perplexity values.
However, this depends strongly on the size of the projected dataset, and on the chosen hyperparameters.
In our guiding example, we projected sorting trajectories for 720 different initial states, which resulted in a total of 8,640 non-unique states (i.e., including duplicates).
With the implementations we used (see Section~\ref{sec:impl}), \tsnet-SNE with perplexity 100 was in fact slightly faster than UMAP with 25 nearest neighbors.
However, for larger datasets, higher dimensionality, or different hyperparameters, UMAP may perform better by up to two orders of magnitude, as reported by Espadoto et al.~\cite{espadoto_towards_2019}.

Note that the state space in the guiding example of sorting algorithms for lists of length 6 is small enough for all possible initial states to be tested easily, and the state space can thus be projected as a whole.
For more complex problems, such as those described in Section~\ref{sec:applications}, the state spaces are much larger, rendering it impossible to project them as a whole.
In these cases, only the subspace of actually visited states is projected.

\subsection{Patterns in Multiple Trajectories}
\label{sec:patterns}

\begin{figure}
    \centering
    \includegraphics[width=\textwidth]{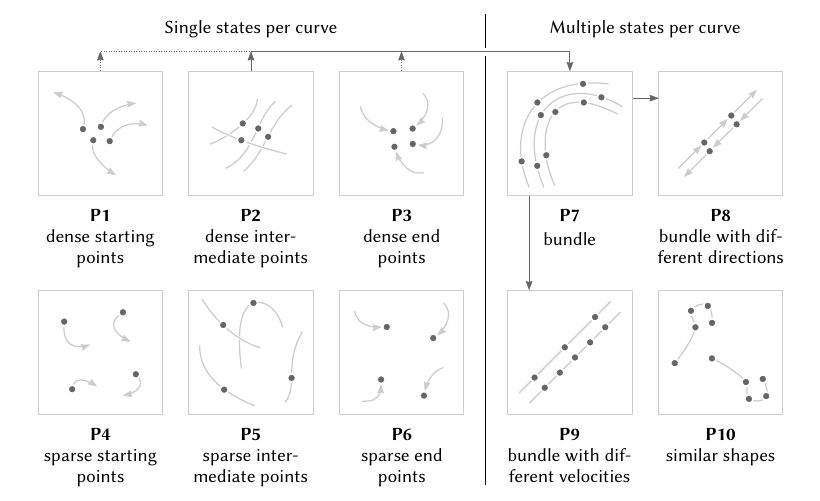}
    \caption{Possible patterns emerging from multiple trajectories. The different kinds of points---starting, intermediate, or end points--of several trajectories can be either sparsely distributed over large regions of the embedding space, or more densely packed.
    This results in six patterns: dense starting points~(\pattern[1]{\label{pat:dense-start}}), dense intermediate points~(\pattern[2]{\label{pat:dense-inter}}), dense end points~(\pattern[3]{\label{pat:dense-end}}), sparse starting points~(\pattern[4]{\label{pat:sparse-start}}), sparse intermediate points~(\pattern[5]{\label{pat:sparse-inter}}), and sparse end points~(\pattern[6]{\label{pat:sparse-end}}).
    Several consecutive intermediate point clusters (possibly together with starting and/or end point clusters) lead to trajectory bundles~(\pattern[7]{\label{pat:bundle}}).
    Trajectory bundles can differ in terms of direction~(\pattern[8]{\label{pat:bundle-dir}}), and/or state space velocity~(\pattern[9]{\label{pat:bundle-velo}}).
    The trajectories can also have similar shapes while populating entirely different regions of the embedding space~(\pattern[10]{\label{pat:shape}}).}
    \Description{Different patterns that can emerge from constructing multiple embedded trajectories together.}
    \label{fig:patterns}
\end{figure}

As can be seen in the guiding example in Figure~\ref{fig:guiding-example}, when multiple trajectories are constructed together, several different patterns can emerge.
Figure~\ref{fig:patterns}\,(a--f) shows the patterns we identified for start, intermediate, or end points, depending on whether they are sparsely distributed over large regions of the embedding space or form a dense cluster.

Obviously, tightly packed starting points~(\ref{pat:dense-start}) can be observed when the visualized processes are constrained to start from the same or from very similar states.
This would be the case, for instance, in games that always start from a fixed initial configuration.
Likewise, many processes converge to the same or to similar end states, which results in a dense cluster of end points~(\ref{pat:dense-end}).
Examples of these processes are list sorting (see Figure~\ref{fig:guiding-example}), optimization routines that reach global optima, and solution paths of puzzles such as the Rubik's cube.
As explained above, in most cases we chose not to remove duplicate entries in the state lists prior to the embedding, which---in the case of \tsnet-SNE or UMAP---leads to a tightly packed cluster rather than a single point in the embedding space even for exactly identical states.

Sparse clouds of starting points~(\ref{pat:sparse-start}) in the embedding space typically arise from randomly distributed initial states.
Certain optimization algorithms, for instance, start with randomly initialized states.
In high-dimensional space, the Euclidean distances between any two random states tend to be similar, with a variance limit of only \(7/120\)~\cite{henry2017}.
Most nonlinear dimensionality reduction techniques based on distances in high-dimensional space tend to spread out many similarly spaced high-dimensional points onto a disk in the embedding space.
Thus, also sparse clouds of end points can emerge if the visualized processes can terminate at many different, but similarly distant, end states.
The list-sorting example from Figure~\ref{fig:guiding-example} is difficult to categorize in terms of patterns of starting points, since---due to the small state space---all possible states were used as starting points.

Perhaps the most interesting pattern among those described in Figure~\ref{fig:patterns}\,(\ref{pat:dense-start}--\ref{pat:sparse-end}) is the densely packed cluster of intermediate points~(\ref{pat:dense-inter}) of multiple trajectories through the embedding space.
This pattern gives rise to \emph{trajectory bundles}~(\ref{pat:bundle})---groups of trajectories that share multiple consecutive clusters of intermediate points.
Trajectory bundles also play an important role in the guiding example in Figure~\ref{fig:guiding-example}, as the two sorting algorithms differ mainly in whether or not distinct bundles are present.
In trajectory data mining, such bundles and similar structures have been described as flocks, convoys, swarms, or gatherings~\cite{zheng_trajectory_2015}.

Figure~\ref{fig:patterns}~(\ref{pat:bundle}) shows how a trajectory bundle results from consecutive clusters of intermediate states (possibly also involving clusters of start and/or end states).
It also shows that the trajectories within a bundle can differ in terms of direction~(\ref{pat:bundle-dir}) and/or velocity~(\ref{pat:bundle-velo}).
However, interpreting the velocity is not straightforward.
As explained in Section~\ref{sec:prelim}, the timestamps (if present) are used only for joining points to form trajectories and not for the projection of states.
This means that two points close to each other in the embedding space do not necessarily need to be close in a temporal sense.
A~direct connection between two points means that the states they represent were visited consecutively, but the viewer might still not know about the \enquote{sampling rate}, that is, the actual time between two states.
In our applications, we deal with indexed states that often do not have a timestamp at all.

\emph{State space velocity} can thus be interpreted as the number of state changes undergone between two non-consecutive points.
Consider two trajectories, \(a = \{a_1, \dots, a_n\}\) and \(b=\{b_1, \dots, b_m\}\), that have two shared intermediate point clusters \(X\) and \(Y\).
Consider further that some \(a_i \in X\) and \(a_{i+1} \in Y\), which means that \(a\) progresses directly from \(X\) to \(Y\).
If now some \(b_j \in X\) and \(b_{j+2} \in Y\), but \(b_{j+1}\) lies somewhere between \(X\) and \(Y\), then \(b\) has a \emph{lower} state-space velocity near state \(j\) than \(a\) has at \(i\), because \(b\) needs more intermediate steps to traverse from \(X\) to \(Y\).
Such a scenario is shown in Figure~\ref{fig:patterns}\,(\ref{pat:bundle-velo}).
A~meaningful comparison of state-space velocities is only possible when two trajectories share at least two intermediate point clusters.

Another pattern that can emerge from plotting multiple trajectories together is shown in Figure~\ref{fig:patterns}\,(\ref{pat:shape}).
Two or more trajectories can have similar overall shapes, but lie in completely different regions of the embedding space.
In the case of nonlinear embedding techniques, such as \tsnet-SNE and UMAP, such patterns must be interpreted with care.
The actual transformation from high-dimensional to low-dimensional manifold may differ considerably for two separate state-space regions.
The only true insight gained from such patterns is that none of the states are similar across different trajectories.
When the shapes of many, well-separated trajectories are very similar (asis the case in the neural network application discussed in Section~\ref{sec:nn}), then a certain similaritybetween the paths in the high-dimensional space---and thusbetween behaviors of the real-world processes---can be assumed.

\section{Interactive Visualization Prototype}
\label{sec:prototype}

To explore the possible patterns described in Section~\ref{sec:patterns}, we implemented an interactive visualization prototype.
In this section, we describe the choices for the visual encoding and the implementation details.
In order to adapt this prototype to new application domains, only minimal changes to one particular part of the encoding are necessary.
While the general visualization principle will be discussed here, the domain-specific changes will be covered with each application scenario in Section~\ref{sec:applications}.

\subsection{Visual Encoding and Interaction}

\pathexplorer{} is an interactive visualization prototype for exploring patterns in collections of projected decision making paths.
The \pathexplorer{} user interface (see Figure~\ref{fig:teaser}) consists of two parts: a side panel with several controls and an interactive plot showing the projected trajectories along with two inset detail views.

In the \pathexplorer{} visualization, all states are represented by plot markers whose positions correspond to the results of some  dimensionality reduction technique.
A~marker can be used to encode additional metadata: Its shape can encode categorical metadata, and its size and color can encode categorical or quantitative metadata.

Similar to Bach et al.~\cite{bach_time_2016}, we use B\'ezier interpolation for connecting projected states to form trajectories.
The improved readability compared to straight lines is especially important, as \pathexplorer{} typically displays many trajectories in a shared embedding space.
The color of the connecting lines of the trajectories can be used to encode additional categorical metadata for each path.

To allow full flexibility and rapid adaptation to new application domains, \pathexplorer{} can visualize data from CSV files with minimal requirements: Only the coordinates of the projected states, the trajectory indices, and the relative ordering of states along trajectories must be given.
\pathexplorer{} supports an unlimited number of additional metadata attributes (i.e.,~additional columns in the CSV files).
Users can choose interactively how these metadata attributes are to be encoded in the different visual channels.
Groups of paths and/or states can be filtered by categorical attributes, and, additionally, paths can be filtered by path length.

Users can hover over state markers to see detailed information in an inset (\enquote{Hover State} in Figure~\ref{fig:teaser}).
Multiple states can be selected by drawing a lasso around them.
A~second detail view shows a \emph{fingerprint} of the selected set of states.
This fingerprint view is typically an adapted version of the detail view for single states, often based on a similarity or difference encoding.
These two insets---the detail view for single states and the fingerprint view for sets of states---are the only parts of the \pathexplorer{} visualization prototype that must be adapted to each application scenario.
We will discuss possible similarity encodings for each application scenario separately in Section~\ref{sec:applications}, and we reflect on the design space of these encodings in Section~\ref{sec:discussion-detail-view}.

Additionally, \pathexplorer{} allows interactive dimensionality reduction using \tsnet-SNE or UMAP.
This feature is helpful if no pre-computed projection has been supplied in the CSV, or if users want to explore different hyperparameters of the embedding techniques.
Users can select which data attributes they want to consider when calculating the embedding.
Since both \tsnet-SNE and UMAP are computationally expensive, the results are shown in a progressively updated animation, which can be terminated at any time.

An additional feature of \pathexplorer{} is the automatic detection of state clusters based on hierarchical density estimates~\cite{hutchison_density-based_2013}.
At the time of writing, this feature is still experimental and requires a local Python backend.

\subsection{Implementation}\label{sec:impl}

The \pathexplorer{} visualization prototype is implemented as a web application written in TypeScript.
We used the three.js framework\footnote{\url{https://threejs.org}} for WebGL rendering of the trajectory plot.
The user interface was implemented using the React library.\footnote{\url{https://reactjs.org/}}
For the interactive dimensionality reduction we used JavaScript implementations of \tsnet-SNE (tsnejs\footnote{\url{https://github.com/karpathy/tsnejs}}) and UMAP (umap-js\footnote{\url{https://github.com/PAIR-code/umap-js}}).
Automatic cluster detection requires a local Python backend and uses the hdbscan\footnote{\url{https://github.com/scikit-learn-contrib/hdbscan}} package.
The \pathexplorer{} prototype can be accessed at
\url{https://jku-vds-lab.at/projection-path-explorer/}.

While we used the \pathexplorer{} prototype for exploring our datasets, the static visualizations in Section~\ref{sec:applications} were created in Python, based on openTSNE results~\cite{policar_opentsne:_2019}. 
For the examples in Figure~\ref{fig:guiding-example} we used openTSNE, the scikit-learn~\cite{pedregosa_scikit-learn:_2011} implementation of Isomap, and the official Python implementation of UMAP by McInnes~\cite{mcinnes_umap:_2018}.

\section{Applications}
\label{sec:applications}

To analyze the usefulness of the patterns described in Section~\ref{sec:patterns}, we applied the \pathexplorer{} visualization protoype (introduced in Section~\ref{sec:prototype}) to high-dimensional processes from three different application domains: Rubik's cube, chess games, and neural network training.
For each application scenario, we first introduce the most important domain-specific concepts, then we describe the state-space representation (cf.~Section~\ref{sec:prelim}), and finally we discuss how the different patterns relate to the real-world processes.%


\subsection{Rubik's Cube Solution Algorithms}
\label{sec:rubik}

Rubik's cube is a famous puzzle toy devised in 1974 by the Hungarian inventor and professor of architecture Ern\H{o} Rubik.
The classic Rubik's cube has six faces, with each face being made up by a \(3 \times 3\) grid of colored facets.
These facets are the faces of smaller cubes, the so-called cubies.
The cube consists of 26 such cubies: 8 corner cubies with three facets each, 12 edge cubies with two facets each, and 6 center cubies with one facet each.
The cube is considered solved when each of its faces shows only one color.

Rubik's cube is commonly known to be almost impossible to solve if no specific solution strategy or algorithm is applied.
The closer a cube is to being solved and the more cubies are in the correct position, the more likely it is that any rotation made with the intention of solving another part of the cube will scramble already correctly placed cubies.
Therefore, precisely considered sequences of rotationsmust be applied which ensure that only specific cubies are moved to their intended destinations.
Many solution strategies for Rubik's cube have been developed.
They differ in complexity and speed, depending on thenumber of special patterns and conditions that are detected and utilizedin the solving process.
Generally, the faster a solution algorithm is and the fewer rotations are needed, the more sub-algorithmsmust be learned and applied under the correct conditions.
The classic beginners' method is highly inefficient, but has only a few sub-algorithms to be memorized.
More advanced methods, such as Fridrich's CFOP method~\cite{Fridrich1997}and the Petrus method~\cite{Petrus1997}, are harder to learn, but usually require significantly fewer moves.
Generally, solution algorithms often use checkpoints: special points in the state space of the cube (e.g., havinga yellow cross on the yellow side).
This state space in which the solution algorithms act is high-dimensional and encompasses more than \(4.3 \times 10^{19}\) unique states.

\subsubsection{State-Space Representation}

The first step of visualizing the solution pathways is to transform the cube states into numerical representations for further processing.
We based the encoding of the cube state on the data structure underlying a Python Rubik's cube  API~\cite{Rubiks-Cube-Solver}, which we also used for calculating the solutions.
Each face of the cube is represented by a \(3 \times 3\) matrix with one entry for each facet.
For the entries representing the facet colors, we chose a one-hot encoding (i.e., (0,0,0,0,0,1) for red, (0,0,0,0,1,0) for green, etc.).
This encoding facilitates the definition of meaningful distance metrics for the dimensionality reduction.
Flattening the resulting \(6 \times (3 \times 3) \times 6\) tensor yielded a feature vector of length 324 for a single state.

Similarity in the original feature space is defined via a distance metric.
We tried different metrics and ultimately chose the Euclidean distance.
Euclidean distance in the high-dimensional state space does not take into account the number of operations (i.e.,~rotations of cube slices) necessary to go from one state to the next.
Instead, we argue that applying the Euclidean distance to our choice of feature vectors yields a representation that is more in line with the intuitive judgment of how scrambled the cube is.
In fact, for this one-hot encoding, the Euclidean distance is equal to the square root of the Hamming distance.
The Euclidean distance thus corresponds to a \enquote{naive edit distance}, that gives a measure of the number of cube facets with the wrong color, but not the number of moves required to fix them.

\subsubsection{Implementation and Visualization Details}

The solutions were calculated from random initial states.
Using a Python Rubik's cube API~\cite{Rubiks-Cube-Solver}, the initial states were generated by performing a set of random rotations on an initially solved cube.
This ensured that only physically possible cube states were used.
We added a new solution algorithm and options for data export to the API.
For each state, we exported~(i)~the high-dimensional encoding and~(ii)~whether that state is a checkpoint of the algorithm used.
Finally, we projected the high-dimensional cube states using \tsnet-SNE.
For the 200 solution trajectories we used a learning rate of 100 and a perplexity of 50.
For the plots with only 2 trajectories we used a perplexity of 20.

The trajectories for different solution algorithms are visually encoded by hue.
The marker shape encodes whether a state is an initial state, a final state, or a checkpoint of the algorithm.
Disk markers for intermediate states between checkpoints can be displayed on demand.
The markers' brightness encodes the progression through the solution trajectory, with bright colors corresponding to early states.

The single-state detail view (see Section~\ref{sec:prototype}) shows the colored facets of an unfolded cube.
In the fingerprint view for multiple selected states, only those cube facets that have the same colors across all selected states are shown as full-sized, colored squares.
For each remaining, non-constant facet, the color values are counted across all selected states.
Each facet is given the color with the highest number of counts, that is, the most prevalent color for that facet.
The non-constant facets are additionally shrunk and made transparent, with both the facet area and the alpha value being proportional to the count of the most prevalent color.
This encoding ensures that constant facets across multiple states can be quickly identified, while information about the non-constant facets is preserved.
An example of this similarity encoding can be seen in Figure~\ref{fig:rubik-selection}.

A~deployed version of our prototype implementation with pre-selected Rubik's cube solution data can be accessed at \url{https://jku-vds-lab.at/projection-path-explorer/?set=rubik}.

\subsubsection{Results }

\begin{figure}
    \includegraphics[width=\textwidth]{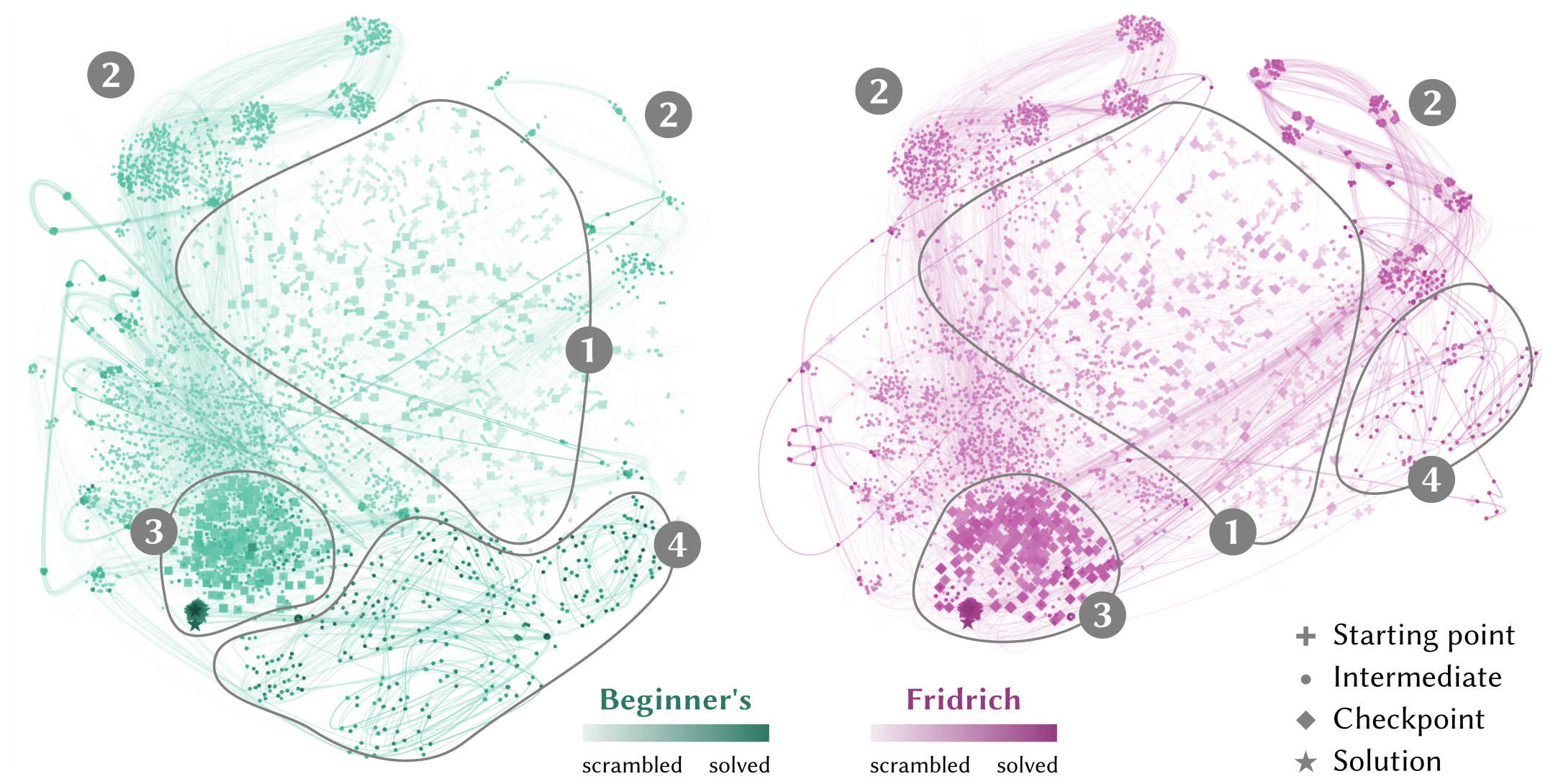}
    \caption{Projected solution pathways for 100 random Rubik's cubes solved with the beginner's method (left), and with Fridrich's method (right), respectively.
    Data for both algorithms was combined for the calculation of the \tsnet-SNE projection, but is shown in two individual visualizations for easier interpretation.
    The random initial states form a broad cluster~(\ref{pat:sparse-start})} near the center of the projected state space~(1).
    Both solution algorithms take similar intermediate paths~(2), and later checkpoints cluster densely~(\ref{pat:dense-inter}) near the final solution~(3).
    Notably, Fridrich's algorithm avoids lengthy sequences of rotations to transform an almost solved cube into a solved cube~(4).
    \Description{Time curve visualization of two Rubik's cube solution algorithms (beginner's and Fridrich method) for random initial states.} 
    \label{fig:rubik-results-overview}
\end{figure}

\begin{figure}[t]
    \includegraphics{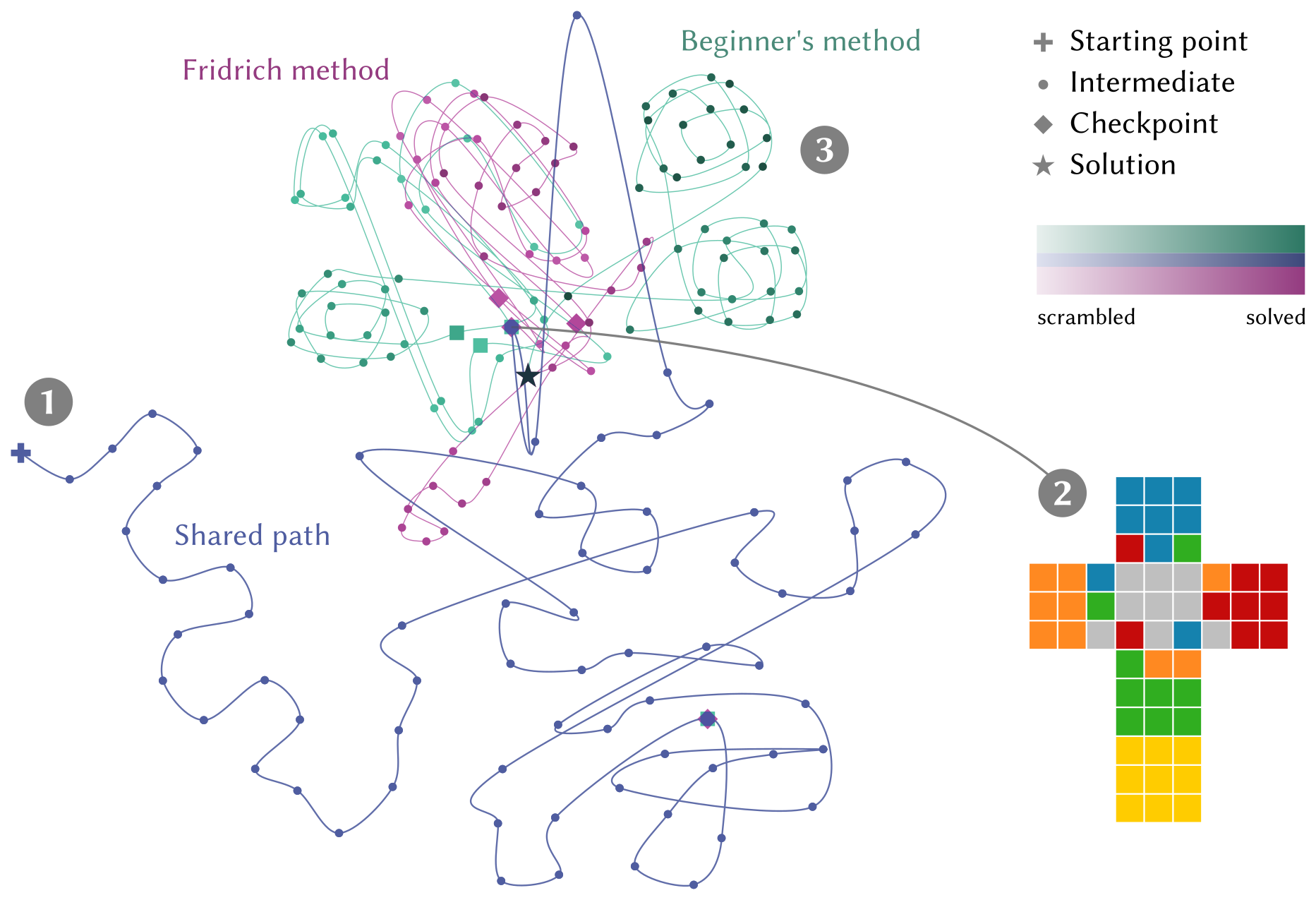}
    \centering
    \caption{Projected solution trajectories of the same initial state (1) solved with the beginner's method and Fridrich's method, respectively.
    The algorithms share the same path up to the second checkpoint (2), at which two layers of the cube are already fully solved.
    Near the end, the beginner's method requires additional rotations, while Fridrich's method approaches the solution much faster.}
    \Description{Trajectory visualization of two Rubik's cube solution algorithms (beginner's and Fridrich method) for equal initial states.}
    \label{fig:rubil-results-detail}
\end{figure}

Figure~\ref{fig:rubik-results-overview} shows the projected solution trajectories for 100 randomly chosen initial cube states, both for the beginner's method and the more advanced Fridrich's method.
Clearly, projecting the cube states with \tsnet-SNE leads to the formation of different clusters.
Most prominently, later checkpoints form dense clusters close to the final solution (see (3) in Figure~\ref{fig:rubik-results-overview}).
Earlier checkpoints with fewer correctly positioned cube facets are much more spread out.
These checkpoints share a wide, sparse region with the randomly selected initial states~(1).
For the checkpoints, we thus observe a correlation between the timestamp \(t_i\) (see Table~\ref{tab:time-curve-definitions}) and the type of pattern: small \(t_i\) correspond to \ref{pat:sparse-inter}, while larger \(t_i\) correspond to \ref{pat:dense-inter}.
In later stages along the solution paths, intermediate states also tend to form dense clusters~(\ref{pat:dense-inter}), which leads to bundles of parallel trajectories~(\ref{pat:bundle}) between checkpoints~(2).



Figure~\ref{fig:rubil-results-detail} shows solution trajectories for the beginner's method and Fridrich's method applied to the same initial state~(1).
Our choice of color-mixing and the use of opacity reveal that the solution trajectories overlap completely up to the second checkpoint~(2).
This almost perfect overlap---a special case of the bundle pattern (\ref{pat:bundle})---shows that even without removing duplicates, \tsnet-SNE does not always introduce significant spacing between identical points for certain hyperparameter choices (in this case perplexity 20).

As, upon user demand, the visualization provides detailed views of unfolded cubes for all intermediate states, it can be seen that two layers of the cube are already fully solved at the second checkpoint.
We found this behavior to be general, that is, independent of the initial state.
Afterwards, the more optimized Fridrich method uses a large number of different sub-algorithms.
This avoids additional lengthy sequences of rotations close to the final solution.
For the beginner's method, these rotations show up as characteristic coils~(3).

When more cube solutions are projected, the differences between the strategies due to the use of sub-algorithms become even more apparent (see Figure~\ref{fig:rubik-results-overview}).
Our visualization shows large bundles of trajectories~(\ref{pat:bundle}) as a result of clusters of similar intermediate points~(\ref{pat:dense-inter}).
Up to a certain point, these bundles appear similar for the two different techniques~(2).

\begin{figure}
    \centering
    \begin{minipage}[b]{0.33\textwidth}
        \centering
        \includegraphics[width=\textwidth]{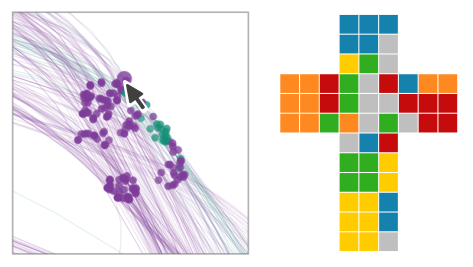}
        \subcaption{Single state}
    \end{minipage}%
    \begin{minipage}[b]{0.33\textwidth}
        \centering
        \includegraphics[width=\textwidth]{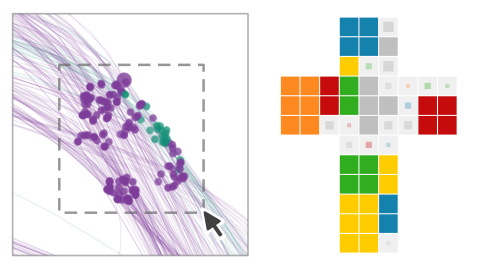}
        \subcaption{Cluster}
    \end{minipage}%
    \begin{minipage}[b]{0.3\textwidth}
        \centering
        \includegraphics[width=\textwidth]{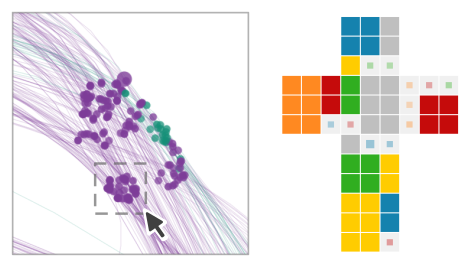}
        \subcaption{Sub-cluster}
    \end{minipage}
    \caption{Analysis of clusters and sub-clusters along trajectory bundles.
    Selecting a single point of a state in a cluster of intermediate points reveals the corresponding cube in the standard detail view~(a).
    Selecting the whole cluster via brushing updates the similarity detail view~(b).
    About 70\,\% of the cube facets are the same across the whole cluster.
    Selecting only the small sub-cluster shows that four more facets are the same (white) inside this sub-cluster~(c).}
    \Description{Example of single and multiple state selection and the similarity encoding for the detail view.}
    \label{fig:rubik-selection}
\end{figure}

{Inspection} of multiple states by means of a lasso selection can help to understand the origin of state clusters and trajectory bundles.
As an example, we look at one state cluster along the trajectory bundle on the upper right-hand side of Figure~\ref{fig:rubik-results-overview}.
The similarity encoding for the cluster is shown in Figure~\ref{fig:rubik-selection}~(b), along with the standard detail view for an individual state in the cluster~(a).
The similarity encoding reveals that about 70\,\% of the facets have the same color across the states in the cluster.
This particular cluster of intermediate points~(\ref{pat:dense-inter}) has a clearly defined sub-cluster, with four more white facets shared across all its states, as seen in Figure~\ref{fig:rubik-selection}~(c).
This more fine-grained analysis is made possible by  the similarity encoding in the fingerprint view.

Further along the solution paths, the beginner's method's inefficient approach to correctly positioning the final cubies gives rise to a large cluster of many \enquote{avoidable} intermediate steps (\ref{pat:sparse-inter}, see (4) in the left part of Figure~\ref{fig:rubik-results-overview}).
In the case of Fridrich's method, many fewer steps are required during the final stages of solving the cube, leading to a much smaller and less populated cluster of intermediate points (see (4) in the right part of Figure~\ref{fig:rubik-results-overview}).

Thus, the scalability of our visualization approach to hundreds of cubes allows us to reliably detect differences between the beginner's method and Fridrich's method.
These differences are most apparent in the different patterns of types \ref{pat:dense-inter} and \ref{pat:sparse-inter}.
Furthermore, the detail views for single and multiple state selections help users understand the structure of the embedding space.
Thereby, these views clarify how the different patterns in the embedding space relate to the real-world cubes.

To tighten this connection between patterns and real-world actions, we built an interactive physical Rubik's cube demonstrator.
This demonstrator combines our visualization approach with a Bluetooth Rubik's cube and a Lego Mindstorms robot.
The demonstrator is described in more detail in the Supplementary Information.


\subsection{Chess Games}
\label{sec:chess}

The game of chess has a compact set of rules.
For each piece---king, queen, rook, bishop, knight, and pawn---only a limited set of movements is allowed.
Nevertheless, chess is extremely rich in tactics and strategy.
Players need to continuously evaluate the positions of the pieces on the board and adapt their future moves accordingly.
To strengthen their skills, players typically study records of games, often with annotations from experts.

A game of chess can be roughly divided into the three stages of opening, middlegame, and endgame.
In the opening phase, players aim to develop their pieces (i.\,e., move them to strategically relevant positions), take control of the center of the board, ensure safety of their king, and structure their pawns.
The middlegame depends mostly on the openings chosen by the two players.
It is typically the phase in which most captures occur, often as results from so-called combinations.
Finally, the endgame is the phase of the game in which only few pieces remain on the board.
Important goals in this stage are promotion of pawns, strategic positioning of the king, and forcing the opponent to make certain moves---a situation called zugzwang.
Players win either by checkmate or resignation of their opponents.
A~game can also end in a draw for a number of reasons, including agreement or a stalemate.

Classic tools for computer-aided studying of chess games, such as Fritz~16~\cite{Fritz16}, typically feature a simple chessboard visualization of one game state at a time.
Instead of focusing on single snapshots of the game, Lu et al.~visualize the evolution of entire games~\cite{lu_chess_2014}.
They introduced the evolution graph: a decision-tree visualization that combines the actual moves performed by the players with alternative moves calculated by AI.
However, this approach remains limited to a single game.

Using our visualization approach, we explore possible similarities between \emph{many} games, and analyze how the games proceed through the different phases.
We apply our visualization approach to chess by viewing the chessboard as a state space, with each configuration of pieces(i.e., each so-called position) corresponding to one possible state.
Each of the players' moves causes the game to proceed from one state to the next.
The number of reachable positions has been estimated to lie between \(2 \times 10^{43}\) and \(1.8 \times 10^{46}\)~\cite{chinchalkar_upper_1996}.

\subsubsection{State-Space Representation}

\begin{table}
    \centering
    \caption{State-space representations for board states before and after various kinds of moves on a simplified 2~\(\times\)~2 chessboard with two different pieces.
    The pieces are encoded as \(\blacksquare\)~=~(0,\,1,\,0) and \(\triangle\)~=~(0,\,0,\,1), and empty fields as (0,\,0,\,0).
    For simplicity of the representation, castling and promotion are performed across colors.
    The listed distances are Euclidean distances with respect to the initial state given in the first row.}
    \begin{tabular}{%
        p{0.15\textwidth}@{}%
        >{\centering\arraybackslash}p{0.11\textwidth}%
        >{\centering\arraybackslash}p{0.19\textwidth}%
        >{\centering\arraybackslash}p{0.29\textwidth}%
        >{\centering\arraybackslash}p{0.11\textwidth}}
        \toprule
        Movement & Board & Tensor & Vector & Distance \\\midrule
        None (initial) & \includegraphics[width=1.2cm, valign=c, padding=0pt .5ex]{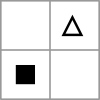} &
        \(\left(\begin{array}{@{\,}c@{~~}c@{\,}}
            (0,0,0) & (0,0,1) \\
            (0,1,0) & (0,0,0) \\
        \end{array}\right)\) & \( (0,0,0,0,0,1,0,1,0,0,0,0) \) & 0 \\
        Simple & \includegraphics[width=1.2cm, valign=c, padding=0pt .5ex]{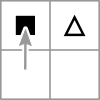} &
        \(\left(\begin{array}{@{\,}c@{~~}c@{\,}}
            (0,1,0) & (0,0,1) \\
            (0,0,0) & (0,0,0) \\
        \end{array}\right)\) & \( (0,1,0,0,0,1,0,0,0,0,0,0) \) & \(\sqrt{2}\) \\
        Capture & \includegraphics[width=1.2cm, valign=c, padding=0pt .5ex]{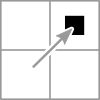} &
        \(\left(\begin{array}{@{\,}c@{~~}c@{\,}}
            (0,0,0) & (0,1,0) \\
            (0,0,0) & (0,0,0) \\
        \end{array}\right)\) & \( (0,0,0,0,1,0,0,0,0,0,0,0) \) & \(\sqrt{3}\) \\
        Castling & \includegraphics[width=1.2cm, valign=c, padding=0pt .5ex]{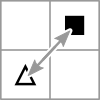} &
        \(\left(\begin{array}{@{\,}c@{~~}c@{\,}}
            (0,0,0) & (0,1,0) \\
            (0,0,1) & (0,0,0) \\
        \end{array}\right)\) & \( (0,0,0,0,1,0,0,0,1,0,0,0) \) & 2 \\
        Promotion & \includegraphics[width=1.2cm, valign=c, padding=0pt .5ex]{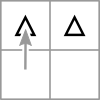} &
        \(\left(\begin{array}{@{\,}c@{~~}c@{\,}}
            (0,0,1) & (0,0,1) \\
            (0,0,0) & (0,0,0) \\
        \end{array}\right)\) & \( (0,0,1,0,0,1,0,0,0,0,0,0) \) & \(\sqrt{2}\) \\
        \bottomrule
    \end{tabular}
    \label{tab:chess-dist}
\end{table}

In order to represent the possible configurations of pieces numerically, we started by viewing the chessboard as an \(8\times8\) matrix.
Each field can be empty or populated by one of 12 different pieces (six different pieces each for black and white).
We thus represented the state of one field by a one-hot code of length 13, which resulted in a \(8 \times 8 \times 13\) tensor encoding the full board state.
Flattening this tensor yieldedtensor a vector of length \(832\).
As in the case of the Rubik's cube visualization, we chose to use the Euclidean distance.
The results of this choice for the different kinds of moves are detailed in Table~\ref{tab:chess-dist}.

Using this encoding and the Euclidean distance has the consequence that moving a bishop diagonally by 1 or 4 fields, for instance, does not result in different state-space distances (as long as no captures occur).
Both resulting states have a distance of \(\sqrt{2}\) to the previous state, and the same is true for all simple movementsand promotions (see first and last row of Table~\ref{tab:chess-dist}).
The states resulting from castling, however, have distance 2 to the previous state, and captures result in a state-space distance of \(\sqrt{3}\).
The type of move (capture, castling, promotion, etc.) leading to each state can additionally be regarded as metadata.

\subsubsection{Implementation and Visualization Details}

The chess game records used for the visualizations in this section are freely available at KingBase~\cite{KingBase} in the Portable Game Notation (PGN) format.
We parsed the PGN files using the chess module of the pgn2gif Python package~\cite{pgn2gif}.
The resulting sequences of chessboard states were encoded as described above.

The trajectories for different opening moves are visually encoded by hue.
Each intermediate state is represented by a marker.
We visualize the initial state (standard chess setup) as crosses, and the final states of the games as stars.
The detail view for a single state selection is simply the chessboard.
The fingerprint for multiple selected states uses a similarity encoding comparable to that used in the Rubik's cube application.
Fields with varying chess pieces across the selected states show only the most prevalent piece.
The piece is made transparent, with an alpha value proportional to the counts for the most prevalent piece on that field across all selected states.

A~deployed version of our prototype implementation with pre-selected chess game data can be accessed at \url{https://jku-vds-lab.at/projection-path-explorer/?set=chess}.


\subsubsection{Results }

\begin{figure}
    \includegraphics[width=0.9\textwidth]{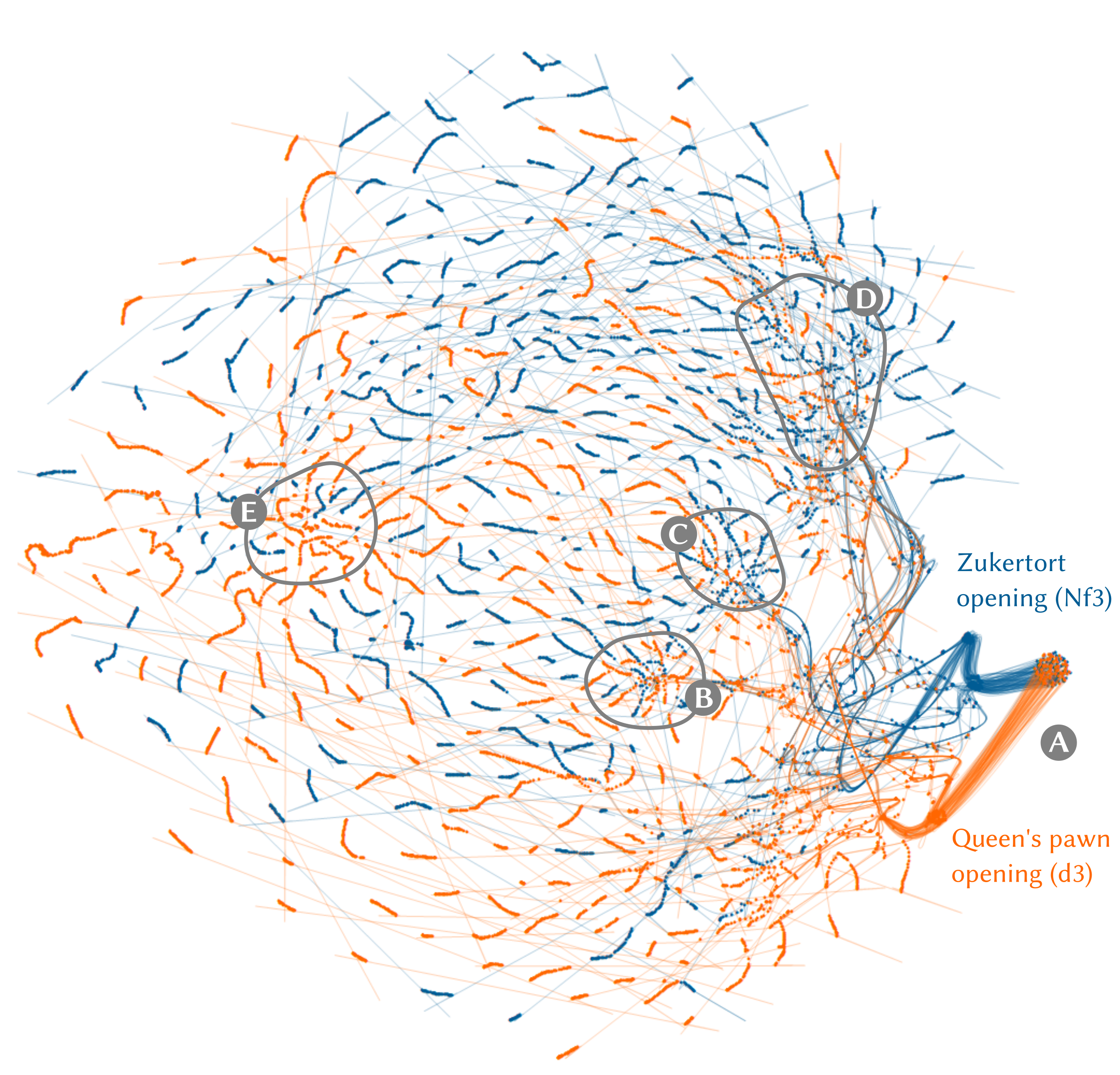}
    \caption{Visualization of 200 chess game trajectories for two different openings, Queen's Pawn Game (d3) and Zukertort Opening (Nf3), respectively.
    Several trajectory bundles~(\ref{pat:bundle}) emerge from the cluster of identical initial states~(a).
    Two smaller clusters of intermediate game states~(\ref{pat:dense-inter}) are visible near the center of the plot.
    Both clusters consist of game states resulting from earlier castling moves.
    The lower cluster~(b) has black pawns placed on c6 and d5.
    The upper cluster~(c) has black pawns placed on c7 and d6, respectively.
    Game states resulting from a certain kind of pawn defense cluster towards the upper right side of the plot~(d).
    Finally, endgames with only few pieces remaining form a more densely packed cluster~(e), while other endgames are spread out evenly across the embedding space~(\ref{pat:sparse-end})}.
    \Description{Trajectories for 200 chess games starting with different moves ((pawn to d3 or knight to f3, respectively).}
    \label{fig:chess-results}
\end{figure}

Figure~\ref{fig:chess-results} shows decision trajectories for 200 randomly selected chess games between players with Elo scores greater than 2,000.
The games are colored depending on the first move by white, which was either the Queen's Pawn Game (d3) or the Zukertort Opening (Nf3).
Together with the King's Pawn Game, these two moves are among the three most common openings.

Since each game starts from the same board state, a dense cluster of initial states~(\ref{pat:dense-start}) is visible in Figure~\ref{fig:chess-results}~(a).
From this cluster, two trajectory bundles~(\ref{pat:bundle}) emerge---one for each of the two openings.
Subsequently, the bundle resulting from the Queen's Pawn Game splits up into one narrow bundle going down and one wide bundle going up.
These bundles reflect the first move performed by black (f5 and e6, respectively).
A~similar splitting can be observed for the games starting with the Zukertort opening.
Even several states later, only a few trajectory bundles are visible in the lower right and right parts of Figure~\ref{fig:chess-results}.
For early timestamps, no sparse intermediate point patterns (\ref{pat:sparse-inter}) are visible.
This makes sense, as for each opening there are only a few established reactions.

The later intermediate states, roughly between the opening phases and the middlegames, form three particular structures that are of type \ref{pat:dense-inter}, but do not give rise to path bundles.
Two clusters, seen in Figure~\ref{fig:chess-results}~(b) and~(c), are collections of game states resulting from earlier castling moves.
The transparency encoding in the interactive detail view revealed that the main difference between these two clusters is the position of two black pawns.
In the lower cluster~(b), two of the black pawns are placed on c6 and d5.
In the upper cluster~(c), they are instead placed on c7 and d6.
Interestingly, games for the two different openings are not represented equally in each cluster.
The Queen's Pawn Game seems to be overrepresented in the cluster related to black pawn positions c6 and d5, while the Zukertort Opening is slightly overrepresented among games in the cluster related to black pawn positions c7 and d6.

Finally, at some point the trajectories tend to make large jumps to a distinct region in the state space where each game culminates in a sequence of densely threaded intermediate states.
These dense sequences (\ref{pat:dense-inter}) before the endstates are representations of endgames.
Endgames in which only few pieces remain converge in a similar region of the state space, as seen in Figure~\ref{fig:chess-results}~(e).
The final states of all other games are sparsely distributed across large regions of the embedding space~(\ref{pat:sparse-end}).

Considering the simple encoding of the chessboard we chose as our state-space representation, it is surprising to see actual gameplay patterns emerge from the embedded trajectories.
These patterns are only possible when many trajectories are constructed together.
Furthermore, the application of decision trajectories to chess games showed that interactive exploration is vital for relating visual patterns to the real-world meaning.
As part of future work, we plan to cooperate with a professional chess player to explore how exactly the patterns correlate with strategic decisions made by players.


\subsection{Neural Network Training}
\label{sec:nn}

Deep neural networks are a class of powerful, nonlinear models that can learn complex representations of data.
They have greatly improved the state of the art in a diverse range of application domains, including computer vision, speech recognition, drug discovery, and genomics~\cite{lecun_deep_2015}.
However, the learning behavior of deep neural networks is difficult to interpret.
It is usually not clear, how a certain choice of hyperparameters or a certain train/test split affect the final performance of the model.
With the growing impact of deep learning on real-world decision-making, these difficulties have led to an increased demand for explainable or interpretable models.
One possible way of analyzing, understanding, and communicating the processes during deep learning is visualization~\cite{hohman_visual_2018}.

The behavior of deep neural networks depends on many different aspects: training and test data, the network architecture, choice of optimization technique and hyperparameters, the actual learned weights, and the resulting activations for each input.
Most of these aspects can be visualized, and almost all of them---even architecture and training data---may change over the course of the training.
As such, the training of neural networks can be viewed as sequential steps through a high-dimensional state space.
Furthermore, developing well-functioning deep models is a highly incremental and iterative process, that requires model builders as well as model users to compare the behavior for many different experimental configurations.
This combination of \enquote{temporality} and the need for comparative analysis makes training of neural networks a perfect candidate for a visualization using multiple embedded trajectories.

While dimensionality reduction has been applied to visualize the representations learned by different network architectures~\cite{mohamed_understanding_2012, hamel_learning_2010,aubry_understanding_2015,donahue_decaf:_2014,mnih_human-level_2015}, in most cases only the projections for one time step are shown---usually for the final network state after termination of the training.
Only few works combine dimensionality reduction techniques with the temporal progression of the training process.
Rauber et al.~visualized the inter-epoch evolution of neuron activations as trajectories in an embedding space~\cite{rauber_visualizing_2017}.
Even more closely related to our work is a visualization used by Erhan et al.~in their study of pre-training of neural  networks~\cite{erhan_why_2010}.
They show learning trajectories through function space, which is one particular state-space representation.
We will show results for two other choices of representations, one of which will also motivate why Erhan et al.~chose their function space representation.

\subsubsection{State Space Representation}

We chose two different state-space representations: a representation of the weight space and a representation of the confusion matrix.

The \emph{weight representation} corresponds directly to the weights learned by the network.
For a given layer, we flatten the associated weight matrix to a single vector.
After each training epoch, the weights are updated and a new weight vector can be obtained.
The length of the vectors, and thus the dimensionality of the state space for this representation, depends on the network architecture.
It is equal to the number of units in that layer, times the number of units in the previous layer.
We use the Euclidean distance as a metric for comparing weight vectors.

The \emph{confusion matrix representation} is more closely connected to network performance.
At each epoch, we let the network classify all test instances, and construct the resulting confusion matrix.
Each cell \((i,j)\) in the confusion matrix lists the number of instances with ground truth class label \(i\) and predicted class label \(j\).
This state-space representation is only suitable for supervised classification problems.
The length of the flat confusion vector is \(k^2\), where \(k\) is the number of different classes.
We experimented with using the Euclidean distance as well as the cosine distance for comparing confusion states, but found that the resulting plots look very similar for both metrics.
Here, we only show trajectories constructed using the Euclidean distance.

\subsubsection{Implementation and Visualization Details}

We used PyTorch (\url{https://pytorch.org/}) to train a simple neural network to classify MNIST~\cite{lecun_mnist_2005} images.
We decided to use a preliminary dimensionality reduction from the number of weights down to 50 components by means of PCA.
According to van der Maaten, this preliminary dimensionality reduction can reduce noise and speed up the computation of the subsequent embedding~\cite{van_der_maaten_visualizing_2008}.
We found that this preprocessing step reduces the computation time significantly, while affecting the plots only negligibly.
We also chose 50 as the number of units in the hidden layer, resulting in 2,500 weights between the input and hidden layers.
We used ReLU as activation function, stochastic gradient descent for optimization, and cross entropy loss.
We tried different learning rates between 0.01 and 1.
In most of our experiments, we trained the networks for 20 epochs, and used the standard MNIST train/test split.

For our experiments with different learning rates, we colored the trajectories categorically by learning rate.
The detail view for single state selection is a visualization of the confusion matrix of the network at this state.
The confusion matrices were evaluated for the MNIST test sets.
Since all networks learned relatively fast, a naive visualization of the confusion matrix would look similar to a diagonal matrix already after the first epoch.
To instead direct the users' attention to the errors, i.e., the off-diagonal values, we left the diagonal blank and adjusted the color scale accordingly.
We recently used a similar approach in a visualization tool for comparing confusion matrices across time~\cite{hinterreiter_confusionflow:_2019}.

A~deployed version of our prototype implementation with pre-selected learning trajectories can be accessed at \url{https://jku-vds-lab.at/projection-path-explorer/?set=neural}.

\subsubsection{Results }

\begin{figure}
    \centering
    \begin{minipage}[b]{0.99\textwidth}
        \centering
        \includegraphics[width=0.95\textwidth]{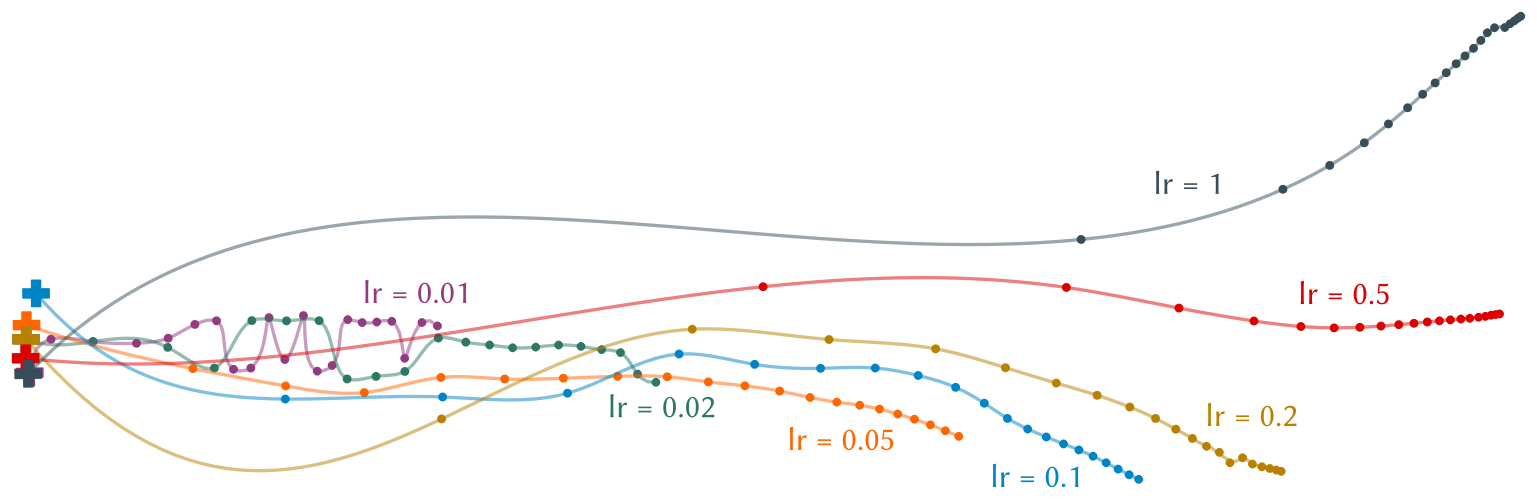}
        \subcaption{Distances between weight matrices of hidden layer}
    \end{minipage}
    \medskip
    
    \begin{minipage}[b]{0.99\textwidth}
        \centering
        \includegraphics[width=.95\textwidth]{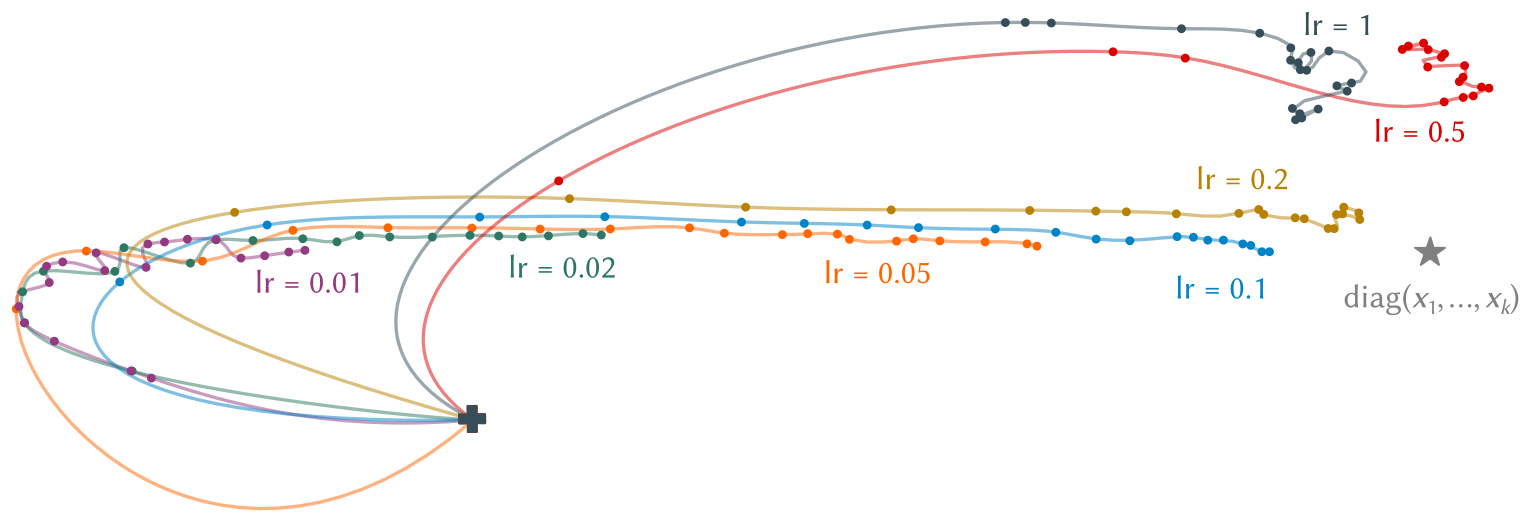}
        \subcaption{Distances between confusion matrices}
    \end{minipage}
    \caption{Learning processes of neural networks trained with different learning rates using a dimensionality reduced (PCA) version of the MNIST images as training data.
    The initialization was the same for all networks.
    Trajectories projected by \tsnet-SNE from two different state-space representations are shown: (a)~based on the weight matrices of the hidden layer; and (b)~on the confusion matrices for the test dataset.
    Both representations resulted in groups of trajectories with similar patterns~(\ref{pat:dense-start}, \ref{pat:bundle}, \ref{pat:bundle-velo}).
    The state space of the confusion matrix representation can be made interpretable by augmenting it with the confusion matrix of perfect classification, denoted by \(\mathrm{diag(x_1, \dots, x_k)}\).}
    \Description{Trajectories for neural network learning with different learning rates, constructed using a weight space representation and a confusion matrix representation, respectively.}
    \label{fig:nn-learning-rate}
\end{figure}

For our first experiments, we wanted to see whether the notion of \enquote{learning speed} is conserved in the embedded trajectories for neural network training.
To this end, we used several different learning rates, and initialized all networks with the same (randomly chosen) initial weights.
The corresponding trajectories are shown in Figure~\ref{fig:nn-learning-rate}, where the top plot~(a) shows trajectories constructed from the weight space representation, while the bottom plot~(b) shows trajectories constructed from the confusion matrix representation.
Different learning rates are encoded by hue, and the values are listed in colored insets.

In both plots, the equal initial states are represented by a cluster of dense  states~(\ref{pat:dense-start}).
It can be seen that the dimensionality reduction technique---in this case \tsnet-SNE with perplexity 40---does not always behave in the same way with regards to clustering equal states, even with equal hyperparameters.
In the weight space representation, the initial states are slightly spread apart, while in the confusion embedding space, they are extremely tightly packed.
The overall trend, however, is the same for both representations.

Lower learning rates lead to smaller changes in the networks' weights, as well as in their classification behavior.
This slow progression through the state space is reflected in the trajectories by a high density of intermediate states, i.e., the learning rate is directly reflected in the different state-space velocities in the path bundle~(\ref{pat:bundle-velo}).
This is also the case near the end of the training, when the optimization starts to converge.

Figure~\ref{fig:nn-learning-rate}~(a) shows that, in the weight space, the trajectories for learning rates between 0.01, 0.2 all move in the same direction.
The trajectories for 0.5 and 1 move into a different region of the embedded state space.
However, from the weight space representation, no insight about performance can be gained.
Even from the confusion embedding, without any additional tricks, only insight about similarity of performances can be gained.
Absolute performance, i.e., overall accuracy, cannot readily be determined from the embedding.
To fix this shortcoming, we added additional structure to the confusion embedding space, by augmenting it with the projected confusion matrix for perfect classification.
This single state was added to all other confusion matrices before performing the dimensionality reduction.
It is shown as a gray star in Figure~\ref{fig:nn-learning-rate}~(b), and denoted by \(\mathrm{diag}(x_1,\dots,x_k)\), since it is a diagonal matrix with the class distribution along its main diagonal.
In this augmented state space it  is possible to interpret intermediate state in terms of network performance, by assessing the distance to the added diagonal matrix.
Calculating the classification accuracy for all learning rates after 20 epochs gives slightly lower values for 0.5 and 1.
A~learning rate of 0.2 yielded the best classifier in terms of accuracy, as would be expected from the embedded trajectories.

\begin{figure}
    \centering
    \begin{minipage}[b]{0.49\textwidth}
        \centering
        \includegraphics[width=0.95\textwidth]{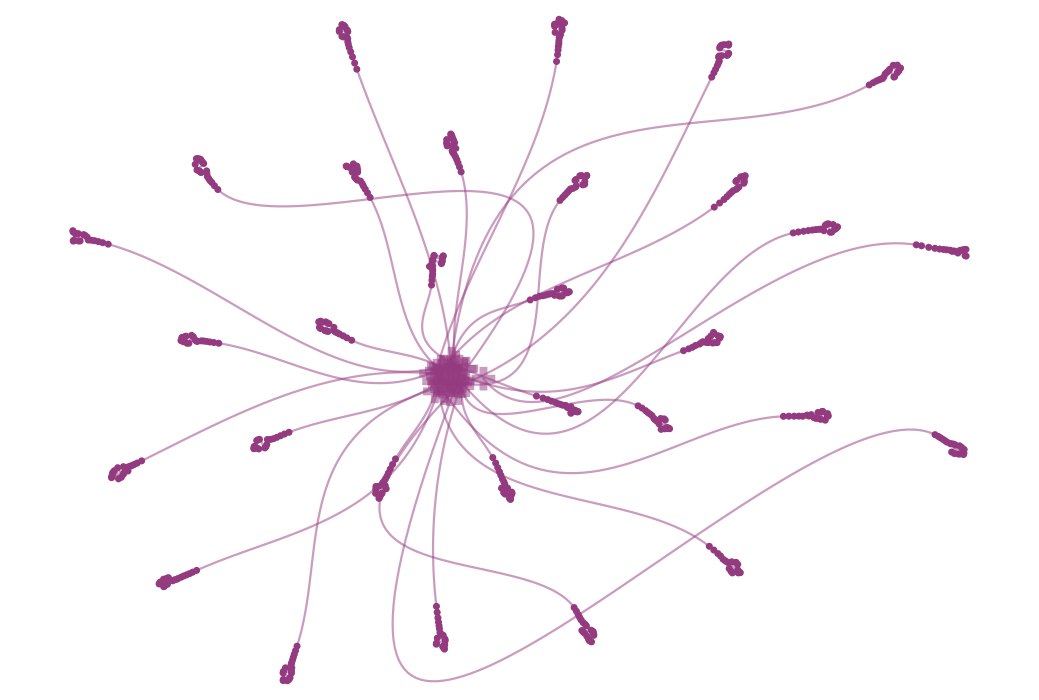}
        \subcaption{Distances between weight matrices of hidden layer}
    \end{minipage}%
    \begin{minipage}[b]{0.49\textwidth}
        \centering
        \includegraphics[width=.95\textwidth]{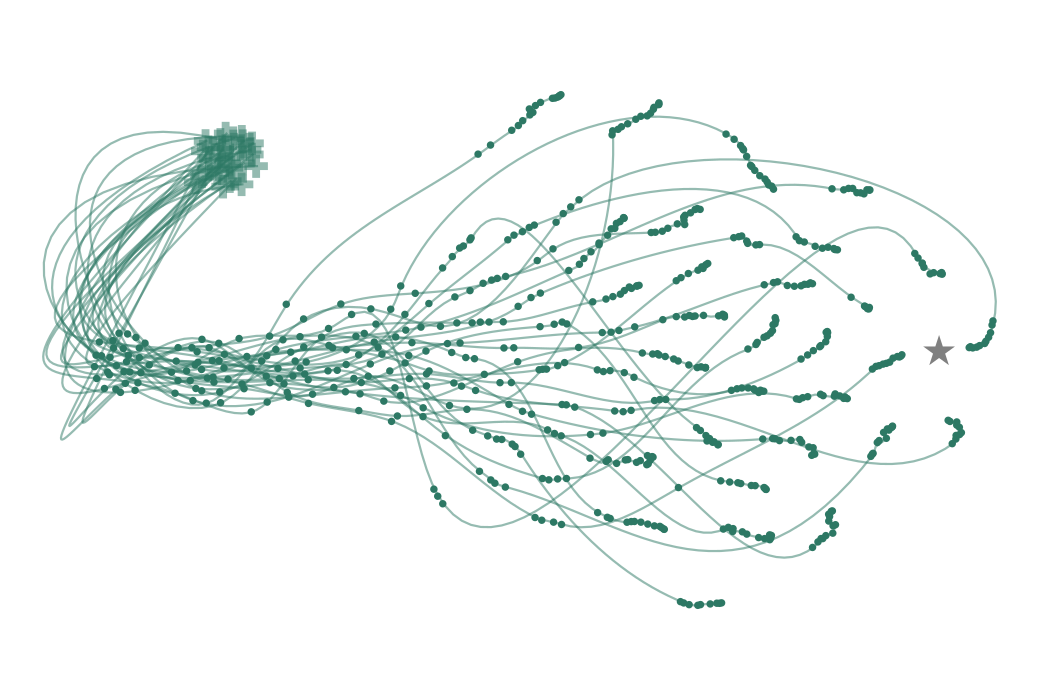}
        \subcaption{Distances between confusion matrices}
    \end{minipage}
    \caption{Learning process of neural networks trained equal learning rates but different random initialization, using a dimensionality reduced (PCA) version of the MNIST images as training data.
    Trajectories embedded using \tsnet-SNE, based on two different state-space representations, are shown: (a)~based on the weight matrices of the hidden layer; and (b)~on the confusion matrices for the test dataset.
    For the confusion matrix representation, the state space is augmented with the confusion matrix of perfect classification, which is denoted by the gray star.}
    \Description{Embedded trajectories for neural network learning with randomly initialized weights, constructed using a weight space representation and a confusion matrix representation, respectively.}
    \label{fig:nn-randinit}
\end{figure}

From Figure~\ref{fig:nn-learning-rate}, one might draw the conclusion that the two representations---based on weights or confusion---carry mostly the same information.
However, this is generally not the case!
Figure~\ref{fig:nn-randinit} shows 30 networks trained on MNIST data, all with the same learning rate (chosen as 0.2 based on the insight from the previous example) and the same architecture.
The only difference between the 30 networks is the random initialization of the weights.
Starting from a dense cluster of initial states~(\ref{pat:dense-start}) in the weight space embedding, all networks quickly move towards their own region, where they converge in a chain of densely packed intermediate states~(\ref{pat:dense-inter}).
Near the end of this chain, almost all trajectories exhibit a similar, bulge-like shape~(\ref{pat:shape}).
This pattern corresponds to the final convergence, when the differences between states become so small that their embedded representations start to form a ball rather than a chain.

The reason for the appearance of similarly shaped trajectories~(\ref{pat:shape}) in different parts of the embedding space---rather than parallel path bundles~(\ref{pat:bundle})---in Figure~\ref{fig:nn-randinit}~(a) is a phenomenon known as weight space symmetries.
Many different weight configurations can result in the exact same mapping function from inputs to outputs.

To address this issue, Erhan et al.~\cite{erhan_why_2010} chose their function space representation.

We chose a confusion matrix representation, which---as detailed above---in addition to revealing differences in the mapping functions also can be interpreted in terms of classification accuracy.
In this embedding, shown in Figure~\ref{fig:nn-randinit}~(b), it becomes clear that the networks are indeed much more similar with respect to their classification and learning behavior as one would expect from the weight space embedding.
Especially in the early learning phases, distinct path bundles form~(\ref{pat:bundle}).
However, regardless of whether the function or the weight space representation is chosen, care must be taken in interpreting the distances in the trajectory visualizations.
In Figure~\ref{fig:nn-randinit}~(b), absolute distances between final confusion matrices for differently initialized weights are artificially \enquote{blown up} by the projection technique trying to conserve relative distances within the chains of very similar matrices near the end of each trajectory.
Nevertheless, this example demonstrates the power of using different state-space representations for showing different aspects of high-dimensional processes with embedded trajectories.

\section{Discussion}
\label{sec:discussion}

\subsection{Patterns in the Embedding Space}

Based on the results described in Section~\ref{sec:applications} and our experience with additional data from Go games and user interactions, we are confident that finding patterns~(\ref{pat:dense-start}--\ref{pat:shape}, as described in Section~\ref{sec:patterns})---in combination with interactive exploration of the detail and fingerprint views---can provide interesting insights.
However, we want to stress the importance of verifying that the patterns found are \enquote{real} and not just the side effects of the dimensionality reduction technique.

In general, patterns which are stable across multiple different embeddings are most useful (here, embeddings can differ in terms of technique or hyperparameters).
In the case of Rubik's cube, for instance, we found the shared trajectory bundles~(\ref{pat:bundle}) of both solution strategies to be stable.
Likewise, the additional \enquote{coils} for lengthy rotation sequences appeared across multiple \tsnet-SNE and UMAP runs (see also Figure~\ref{fig:teaser}, which shows a different embedding than Figure~\ref{fig:rubik-results-overview}).

Generally speaking, we found the dense point-cluster patterns~(\ref{pat:dense-start}--\ref{pat:dense-end}) and the resulting path bundle-patterns~(\ref{pat:bundle}) to be most reliable across all application scenarios.
Sparse patterns~(\ref{pat:sparse-start}--\ref{pat:sparse-end}) are harder to relate to the real-world processes, as they cover many, often considerably different, high-dimensional states.
Only when the association with the real-world process is straightforward (e.g., with random initialization of algorithms), have we found these patterns to be useful anchor points in the analysis.

In the case of the pattern \ref{pat:bundle-velo} in the neural network example (Figure~\ref{fig:nn-learning-rate} in Section~\ref{sec:nn}), we were surprised how well the state-space velocity correlated with the \enquote{true} velocity (i.e., the learning rate).
We think that this pattern has considerable potential in exploring processes that converge with different speeds.

Finally, the hypothetical pattern~\ref{pat:bundle-dir}, which would arise from trajectory bundles with different directions, could not be found in any of the examples presented.
However, we are currently investigating user-interaction data and expect to find such patterns when users backtrack to revisit earlier parts of their analysis.

\subsection{ Dimensionality Reduction}
\label{sec:discussion-dimred}

Clearly, when high-dimensional states are represented in a two-dimensional embedding space, it is impossible to preserve all original distance information.
Additionally, hyperparameters of the projection techniques must be set appropriately.
Depending on the dimensionality reduction technique, several challenges arise.

\begin{figure}
    \centering
    \begin{minipage}[b]{0.49\textwidth}
        \centering
        \includegraphics[width=0.96\textwidth]{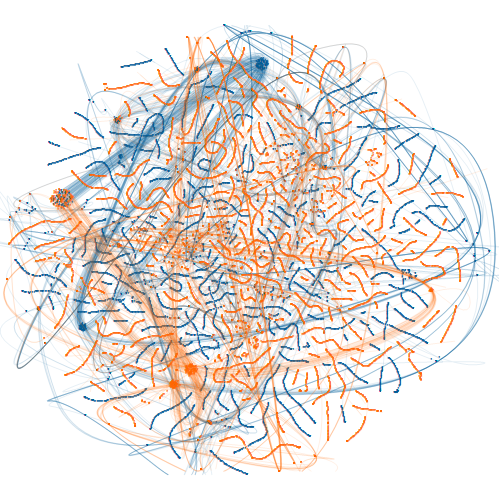}
        \subcaption{\tsnet-SNE, perplexity 5}
    \end{minipage}%
    \begin{minipage}[b]{0.49\textwidth}
        \centering
        \includegraphics[width=.96\textwidth]{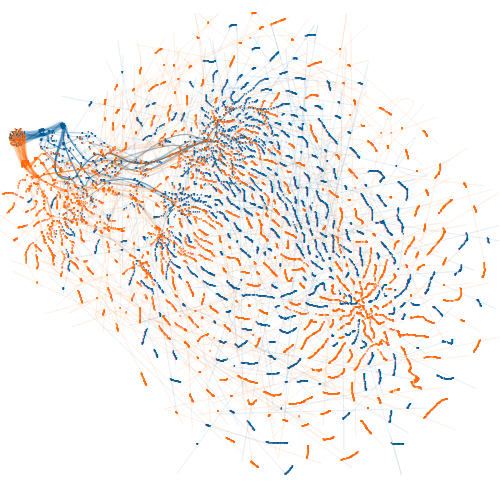}
        \subcaption{\tsnet-SNE, perplexity 300}
    \end{minipage}
    \caption{Embeddings (\tsnet-SNE) of chess games from Figure~\ref{fig:chess-results}, re-run with different perplexity values.}
    \Description{Re-run embeddings of chess games from Figure~\ref{fig:chess-results}, with different perplexity values.}
    \label{fig:chess-stability}
\end{figure}

The most obvious challenge when using \tsnet-SNE is to set the perplexity.
Figure~\ref{fig:chess-stability} shows the same data as Figure~\ref{fig:chess-results}, but embedded with perplexity values of~(a)~5, and~(b)~3000.
Two types of patterns are stable across all perplexity values: the presence of path bundles at the early games stages and the chains of states towards the ends.
However, in the case of perplexity~5 the path bundles appear more \enquote{messy}, and regions of more densely packed intermediate states (as in Figure~\ref{fig:chess-results}\,(b--e)) cannot be made out.
Both of these differences are a result of using a perplexity value that was too low to preserve global structure.

While van der Maaten stated that \enquote{typical values for the perplexity range between 5 and 50}~\cite{tsne-faq}, we generally found higher perplexity values to be more useful for our purposes.
Okolkov suggested a square-root law, that is, choosing a perplexity of \(\sqrt{N}\) for a dataset with \(N\) points~\cite{umap-vs-tsne}.
We suggest experimenting with even higher perplexity values.
Plots often remain relatively stable in a wide perplexity range above \(\sqrt{N}\).
Higher perplexity values work around the fact that the objective function of \tsnet-SNE, as compared to that of UMAP, does not strictly penalize for two points with large high-dimensional distance to be embedded close together.

Another challenge related to \tsnet-SNE is that the result heavily depends on the initialization.
In the \tsnet-SNE implementation that we used~\cite{policar_opentsne:_2019}, random initialization of low-dimensional positions can be replaced by initialization with PCA results.
In our experience, this PCA initialization greatly improves comparability of multiple embeddings, even across different perplexity values.

Occasionally, we experienced some unexpected behavior with \tsnet-SNE: We noticed significant \enquote{jumps} between similar states.
While \tsnet-SNE is known for sometimes putting points of high-dimensionally large distance close to each other in the embedding space (see above), it should not put high-dimensionally close points very far apart in the embedding.
In experiments with user interaction-data, we noticed that UMAP did not produce these jumps as often.
However, UMAP introduces other challenges related to setting the minimal-distance parameter such that interesting patterns related to dense clusters are not hidden.

\subsection{ State Space Representations and Distance Metrics}

Choosing a meaningful combination of state-space representation and distance metric for constructing decision trajectories can be challenging.
As already hinted at in Section~\ref{sec:rubik} about Rubik's cube solution strategies, it might not always be possible to find a state-space representation that can accurately preserve the real-world notion of distances between states.
For the Rubik's cube example, a combination of representation and metric would be ideal if it put two cube states at a distance of 1 if one state could be reached from the other in a single quarter-turn of a cube face.
Even if such a representation existed, it would have to be based on finding the shortest path between two cube states in terms of rotations.
An algorithm capable of solving Rubik's cube with the shortest possible sequence of moves has been termed God's Algorithm, and proving the lower bound of necessary moves alone took 35 CPU-years in 2010~\cite{gods_number}.
To the best of our knowledge, there is still no efficient implementation for finding the shortest path between two arbitrary states.
Therefore, the Rubik's cube visualization shows that sometimes a simpler state-space representationmust be used in the hope that the representation and metric preserve interesting features of the real-world processes.

Furthermore, for many applications certain symmetries could be exploited in the state space.
In the case of the Rubik's cube, many cubes are equivalent with regards to solutions involving, for instance, the yellow or green cross as a first checkpoint.
Likewise, chessboard states could be seen as equivalent if the white and black pieces are swapped.
For chess, it could be argued that this symmetry is broken by the rule of white always performing the first move.
Nonetheless, reflection-symmetrical states could be regarded as equivalent.
In case of Rubik's cube, we broke the symmetries consciously by forcing each cube to pass the same checkpoint (yellow cross)---essentially disregarding more efficient solutions involving different checkpoints.

For both the chess and the Rubik's cube visualizations, disregarding the symmetry still led to interpretable trajectory patterns.
However, many applications could benefit from a state-space representation and/or distance metric that takes symmetries into account.
In recent experiments with Go game data, we found that only symmetry-preserving distance metrics based on wavelet or Fourier transforms of the Go boards led to interpretable visualizations.

\subsection{Construction of Trajectories}

In order to keep visualizations with hundreds of trajectories readable, interpolation between points is extremely important.
However, each interpolation technique may introduce artifacts, such as overshooting splines.
An example of such artifacts can be seen close to the final chain of points for each game in Figure~\ref{fig:chess-results}.

Additionally, not all interpolation techniques are equally suited for adding points to trajectories on the fly, which is necessary when our visualization technique is used for streaming data.
Newly added points may change the shapes of trajectories through many of the previous points considerably.
For our interactive physical Rubik's cube demonstrator (see Supplementary Information), we used cubic cardinal splines to make this addition as unobtrusive as possible.
This spline interpolation technique makes sure that adding a point alters only the trajectory through the last three intermediate points. 

Even with a suitably chosen interpolation technique, trajectories through the embedded state space may cause visual clutter.
We plan to experiment with edge-bundling techniques to address this issue in future versions of our applications of decision trajectories.
In addition to path bundling, smoothing of trajectories (e.g., by 1D Laplacian filtering) could help to reduce this clutter.

\subsection{Detail View and Fingerprint Encoding}
\label{sec:discussion-detail-view}

We mentioned in Section~\ref{sec:prototype} that adapting the \pathexplorer{} visualization prototype to new application domains requires only designing appropriate detail and fingerprint views.
For most real-world processes to which we have applied \pathexplorer{} so far, choosing a detail view for the single state selection was relatively straightforward.
However, the design of an appropriate fingerprint view for multiple state selections poses a greater challenge.
Should the fingerprint focus on similarities or differences between the selected states?
And how can similarities or differences be encoded in a way that intuitively extends the detail view for single states?

In the cases of Rubik's cube and chess, we were able to extend the single-state views by transparency and/or size encodings of shared attributes across selected clusters.
For the neural network data, we did not implement any fingerprint encoding, but it seems that a difference encoding would be more informative in this case.
Finally, in our experiments with user-interaction data, the high-dimensional states are combinations of multiple categorical and numerical attributes, together with representations of sets.
These heterogeneous data types are all combined into mixed-type feature vectors that are then projected.
Designing effective fingerprint encodings for such complex cases, requires a more in-depth study of the design space of similarity and/or difference encodings.

\subsection{Future Work}

As stated in Section~\ref{sec:chess}, we plan to examine the correlation between players' strategies and patterns in trajectories of chess games in more detail.
To this end, we are currently initiating a collaboration with a professional chess player.
We will apply our visualization technique not only to chess games by human players, but also to games played by chess computer programs.
removed{Especially, it}It would be particularly interesting to compare strategies used in classic chess computer programs with those learned by state-of-the art reinforcement learning algorithms such as AlphaZero~\cite{silver_general_2018}.

In light of potential future collaborations with domain experts, we plan to further improve the interactivity of our \pathexplorer{} prototype.
We have already started to implement automatic detection of state clusters (\ref{pat:dense-start}--\ref{pat:dense-end}) based on hierarchical density estimates~\cite{hutchison_density-based_2013}.
We plan to use the detected clusters to draw the user's attention to specific regions in the embedding space. By ranking the clusters according to a degree-of-interest function, we plan to automatically create insets as demonstrated in a recent work by Lekschas et al.~on patter-driven navigation~\cite{lekschas_pattern-driven_2019}.
These insets will make use of the fingerprint view described in Section~\ref{sec:prototype}.
Furthermore, building on techniques from trajectory data mining~\cite{zheng_trajectory_2015}, the detected clusters can be used to automatically find interesting patterns related to trajectory bundles (\ref{pat:bundle}--\ref{pat:bundle-velo}).

We are also currently working on a project aimed at finding and understanding user actions in interactive visualizations.
We plan to employ a similar, but less domain-specific, approach as Brown et al.~did in ModelSpace~\cite{brown_modelspace:_2018}.
Our goal is to extract interaction data from provenance graphs of visualization applications, and to visualize the data as multiple trajectories through the same embedding space.
This embedding space will be based on a representation of the visualization's properties that are changed by  user interactions.
We hypothesize that patterns emerging from multiple projected interaction trajectories can provide insights into the users' sensemaking processes.

\section{Conclusions}
\label{sec:conclusion}

In this paper, we have explored visual patterns in projected problem-solution paths.
This was made possible by viewing the decisions---whether made by humans or resulting from the rules of an algorithm---as transitions between states in a high-dimensional representation space.

To reveal and explore the patterns, we projected multiple paths through the high-dimensional state space as trajectories in a shared, low-dimensional embedding space.
We found this approach sufficiently general to be applied in various application domains: Rubik's cube solution algorithms, Chess games, and neural network training.

In all our applications, we used our interactive visualization prototype, \pathexplorer{}, to analyze the embedded trajectories.
This general prototype can be adapted easily to new domains by defining new encodings for the detail view and the fingerprint encoding for state clusters.
In the future, we plan to adapt the prototype for visualizing user interactions in visual analytics applications.

We hope that our work will inspire further studies of projected decision trajectories in a variety of interesting application domains.

\begin{acks}
This work was supported in part by the State of Upper Austria and the Austrian Federal Ministry of Education, Science and Research via the LIT -- Linz Institute of Technology (LIT-2019-7-SEE-117), by the State of Upper Austria (Human-Interpretable Machine Learning), by the Austrian Research Promotion Agency (FFG851460), and by the Austrian Science Fund (FWF P27975-NBL).
\end{acks}

\bibliographystyle{ACM-Reference-Format}
\bibliography{rubik-extended-papers,rubik-extended-online}

\end{document}